\documentclass[journal]{IEEEtran} 
\usepackage{amssymb}
\usepackage{graphicx}
\usepackage{textcomp}
\usepackage{amsmath}
\usepackage{color}
\usepackage{lineno}
\usepackage{url}
\usepackage{cite}
\usepackage{enumerate}
\usepackage{epstopdf}

\usepackage{graphicx}
\usepackage{graphics}
\usepackage{multirow}
\usepackage{caption}
\usepackage{subcaption}
\usepackage{gensymb}
\usepackage{float}
\usepackage{verbatim}

\usepackage{multirow}
\usepackage{array}
\usepackage{booktabs}
\usepackage{hhline}
\usepackage{xspace}
\usepackage{breqn}
\usepackage{tablefootnote}

\usepackage{algorithm}
\usepackage{algpseudocode}
\usepackage{soul}

\newcommand{\comst}[1]{}

\definecolor{orange}{rgb}{1,0.5,0}

\newcommand{\ours}{Future-State Predicting LSTM }

\newcommand{\com}[1] {}

\graphicspath{{./figures/}}
\allowdisplaybreaks
\usepackage{array}
\usepackage{arydshln}
\usepackage{hyperref}

\begin{document}

\title{Future-State Predicting LSTM for Early Surgery Type Recognition}

\author{Siddharth Kannan, Gaurav Yengera, Didier Mutter, Jacques Marescaux, Nicolas Padoy*

\thanks{This work was supported by French state funds managed within the Investissements d'Avenir program by the ANR (references ANR-11-LABX-0004 and ANR-10-IAHU-02) and by BPI France (project CONDOR). The authors would also like to acknowledge the support of NVIDIA with the donation of a GPU used in this research. \textit{Asterisk indicates corresponding author.}

Siddharth Kannan, Gaurav Yengera and Nicolas Padoy are with ICube, University of Strasbourg, CNRS, IHU Strasbourg, France (email: sid2496@gmail.com, g.yengera@gmail.com, npadoy@unistra.fr).

Didier Mutter and Jacques Marescaux are with the University Hospital of Strasbourg, IRCAD, IHU Strasbourg, France.

\copyright \space 2019 IEEE. Personal use of this material is permitted.  Permission from IEEE must be obtained for all other uses, in any current or future media, including reprinting/republishing this material for advertising or promotional purposes, creating new collective works, for resale or redistribution to servers or lists, or reuse of any copyrighted component of this work in other works.

The final version of this paper is S. Kannan, G. Yengera, D. Mutter, J. Marescaux and N. Padoy, "Future-State Predicting LSTM for Early Surgery Type Recognition" in IEEE Transactions on Medical Imaging. DOI: 10.1109/TMI.2019.2931158. Available at \href{http://dx.doi.org/10.1109/TMI.2019.2931158}{http://dx.doi.org/10.1109/TMI.2019.2931158}
}}%

\markboth{}
{S. Kannan \MakeLowercase{\textit{et al.}}: {\ours} for Early Recognition of Surgery Type}
\maketitle
\begin{abstract}
This work presents a novel approach for the early recognition of the type of a laparoscopic surgery from its video. Early recognition algorithms can be beneficial to the development of 'smart' OR systems that can provide automatic context-aware assistance, and also enable quick database indexing. The task is however ridden with challenges specific to videos belonging to the domain of laparoscopy, such as high visual similarity across surgeries and large variations in video durations. 
To capture the spatio-temporal dependencies in these videos, we choose as our model a combination of a Convolutional Neural Network (CNN) and Long Short-Term Memory (LSTM) network. We then propose two complementary approaches for improving early recognition performance. The first approach is a CNN fine-tuning method that encourages surgeries to be distinguished based on the initial frames of laparoscopic videos. The second approach, referred to as '{\ours}', trains an LSTM to predict information related to future frames, which helps in distinguishing between the different types of surgeries. We evaluate our approaches on a large dataset of 425 laparoscopic videos containing 9 types of surgeries (\textit{Laparo425}), and achieve on average an accuracy of 75\% having observed only the first 10 minutes of a surgery. These results are quite promising from a practical standpoint and also encouraging for other types of image-guided surgeries.
\end{abstract}

\begin{IEEEkeywords}
Laparoscopic Video, Early Detection, Surgery Recognition, Surgical Workflow, Deep Learning.
\end{IEEEkeywords}

\section{Introduction}\label{sec:intro}

\IEEEPARstart{T}{}he advent of Laparoscopic surgery, in addition to having improved patient safety and recovery time, has led to the development of intelligent algorithms relying on laparoscopic video data. Algorithms have been proposed for automatic surgical phase recognition \cite{yengera2018}, remaining surgery duration prediction \cite{twinanda2018rsd}, surgical tool detection \cite{hajj2017}, which are crucial for the development of context-aware systems that can find applications in automatic surgery monitoring, surgical education and human-machine interaction within the OR. Information about the surgery type is important for context-aware systems as it enables the system to perform surgery specific analyses and consequently to deploy a contextual user-interface in the OR that is adapted to the needs of the particular surgery.

Early surgery type recognition approaches can help in the realization of a real-time fully-automated context-aware system by enabling information about the surgery to be automatically obtained. This reduces the dependence on manual input, which can cause disruption in the surgical workflow, and on proprietary systems, in which this information may be available, but which are not always easily accessible. Further applications include quick database indexing and, in general, the proposed approaches could be beneficial for early recognition tasks in other domains of computer vision as well. 

Previous approaches to surgery recognition \cite{twinanda2014} required the complete video to be available, making them more data intensive and restricted to offline use. To the best of our knowledge, this is the first work that focuses on early recognition of the surgery type using laparoscopic videos. There are several unique challenges associated with this task. Firstly, laparoscopic video frames of different surgeries are very similar in appearance leading to low inter-class variability. This is unlike benchmark video classification datasets in the computer vision literature, where different classes have very different contexts. This makes it challenging to capture discriminative information between different classes (surgery types). Secondly, there is high intra-class variability since the duration of surgeries belonging to the same class can vary over a wide range due to differing patient conditions and surgeon skill levels. Additionally, we also have to cope with blood-stained camera lenses, motion blur, presence of smoke and deformations due to tool-tissue interactions.

In this work, we aim to capture the spatial characteristics of laparoscopic video data by utilizing a Convolutional Neural Network (CNN) and the temporal information as the surgery evolves by using a Long Short-Term Memory (LSTM) network. Our focus is on presenting approaches to train the CNN and LSTM networks for improving the early recognition performance. To this end, we present a method for CNN fine-tuning that encourages the network to learn discriminative features from the initial frames of surgical videos. Further, we develop a novel approach where we train an LSTM model, referred to as "{\ours}" (FSP-LSTM), to anticipate the future hidden states of a previously trained LSTM model, the teacher network, in addition to predicting the type of surgery. This enables the knowledge accumulated by the teacher LSTM over time to be imparted to the FSP-LSTM. We believe that this enables the FSP-LSTM to correlate present video frames with expected future frames ahead of time, and thereby helps improve early surgery recognition performance. We also explore variations in the loss function, as done in prior work on early action recognition \cite{aliakbarian2017iccv}.

The rest of this paper is organized as follows: Section \ref{sec:related_work} discusses prior work on video classification, surgery classification, future prediction and action anticipation. Section \ref{sec:method} details the developed approaches. The laparoscopic dataset and experimental setup are described in section \ref{sec:exp_setup}. A study of recognition performance at different time steps is presented in section \ref{sec:results}. Section \ref{sec:discussion} presents an ablation study and further discussion of the proposed models, which is followed by a conclusion in section \ref{sec:conclusion}.

\section{Related Work}\label{sec:related_work}

\subsection{Video Classification}
\label{sec:related_work_vclassif}
Video classification has been a challenging problem in the computer vision community because of the need to capture both spatial and temporal information. Traditional methods have used hand-crafted features like Histogram-of-Oriented-Gradients (HOG) and Scale-Invariant-Feature-Transform (SIFT) for extracting spatial information. These features are then aggregated using encoding schemes like Fisher Encoding or K-Means clustering to build a Bag-of-Visual-Words \cite{laptev2008cvpr} to produce video-level predictions. Methods like Improved Dense Trajectories \cite{wang2013iccv}, which explicitly model motion information in successive video frames making use of Motion-Boundary-Histograms and Optical-Flow, had previously achieved state-of-the-art results on standard video recognition datasets like HMDB51 and UCF-101.

More recently, deep learning architectures, like CNNs, have advanced from strength to strength to achieve state-of-the-art performance, almost on par with humans, on image recognition tasks \cite{karpathy2012,simonyan_vgg14,he2016}. This has in turn encouraged attempts to apply them to video recognition tasks as well. However, they are not as effective when it comes to videos due to their inability to extract temporal patterns that are crucial for learning video representations. In order to address this, Karpathy et al. \cite{karpathy2012} extended the spatial connectivity of convolution kernels to the time-domain and thereby perform 3D-convolutions to extract spatio-temporal information from short video clips. But this method improves marginally over the standard CNNs trained on single frames and moreover, may not be feasible for videos of long durations due to memory constraints. Simonyan et al. \cite{simonyan2014} proposed a two-stream CNN architecture to learn appearance and motion features and fuse them to make a prediction. They achieved state-of-the-art results on the UCF-101 dataset. They also report better performance from using only optical-flow features as compared to using only RGB features and thereby demonstrate the importance of leveraging temporal information for learning video representations. But optical-flow computation can be expensive and therefore, this approach may not be suitable for the real-time nature of our task. The above approaches also have the limitation that a fixed number of frames have to be used as input to the CNN, thus, favoring the learning of shorter videos.

Ng et al. \cite{ng2015} suggest two approaches for capturing temporal information from variable length temporal data. In the first, they explore pooling strategies at different stages of the CNN, which are agnostic to the size of the dimension they operate on, for aggregation of features along the temporal dimension. They report significant improvements over \cite{karpathy2012} on the Sports-1M dataset by max-pooling features extracted from longer video clips that span most of the video and show that their model benefits from using global information from the video as opposed to using frames from a localized region. In particular, this can be effective for laparoscopic videos which are of long durations and contain defining surgical events occurring at various points of time in the video. But unlike \cite{ng2015}, we are limited by the fact that our videos have an average duration of 86 minutes and therefore, cannot use all the frames for training the CNN proposed in their first method due to memory constraints. Instead, we show that training a CNN on a sparse uniform sampling of just 1\% of the video is also an effective way to gather global information and yields an improved performance as compared to training a CNN on just individual frames. The second approach of \cite{ng2015} uses LSTM cells to extract temporal information from the features computed by the CNN that precedes it. The advantage of using an LSTM is that it can capture long-term temporal dependencies and is also known to avoid the vanishing gradient-problem that traditional recurrent networks are susceptible to \cite{hochreiter1997long}. Additionally, the LSTM can make use of the order of occurrence of frames which the aforementioned pooling strategies fail to take advantage of and consequently, yields the best performance for \cite{ng2015} on UCF-101, although beating their first approach only by a narrow margin.

\subsection{Surgery Classification}
While the aforementioned methods on video classification have achieved state-of-the-art performance on benchmark video recognition datasets like UCF-101 and Sports-1M, work on surgical video recognition has received little attention primarily due to the dearth of large-scale datasets in this domain. Previous works in surgical video analysis have addressed problems like surgical phase recognition \cite{twinanda2017endonet}, surgical gesture classification \cite{zappella2012miccai}, tool tracking \cite{reiter2012miccai}, classification of anatomical structures and surgical actions from video shots \cite{petscharnig2018}. The most relevant to this work is the paper by Twinanda et al. \cite{twinanda2014} in which they propose a pipeline for classification of the type of Laparoscopic video, which consists of frame rejection, feature extraction, feature encoding and classification. They report an accuracy of 90.38\% on a dataset of 208 laparoscopic videos, divided into 8 classes, using the generalized multiple kernel learning (GMKL) algorithm \cite{varma2009}, which combines the best encoding kernels for RGB histograms, hue-saturation histograms, SIFT and HOG respectively. However, the extraction of hand-crafted features can be computationally intensive and hence, not suitable for our application. Moreover, deep features have been reported to outperform hand-crafted features for recognition tasks on Laparoscopic Videos in both online and offline setups \cite{twinanda2017endonet}. Hence, we propose to use deep network architectures in our work. Twinanda et al. \cite{twinanda2014} also perform sub-video classification with their proposed pipeline so as to use a minimal number of frames for accurate classification of the videos. The reported accuracy is over 70\% using only the last 20\% of each video, while the accuracies are lower when only the first 20\% of the videos are used. This suggests that the most discriminative part of the video is towards the end. Given that the dataset we use for this work has the same classes as in \cite{twinanda2014} and one additional class, the above-mentioned observations have implications on the feasibility of our task using such methods since we wish to classify surgeries as early as possible. We hope to overcome this potential challenge by encouraging our model to gather discriminative information from the initial frames of laparoscopic videos and also by leveraging the novel loss functions proposed in previous works on action anticipation \cite{jain2016}, \cite{aliakbarian2017iccv} to encourage a model to learn the early recognition of the classes.

\subsection{Future Prediction of Visual Representations}
Prior work on future prediction related to our approach have involved prediction of future frames of a video at the pixel level \cite{ranzato2014, srivastava2015, lotter2017} or anticipation of high-level visual representations \cite{vondrick2016}. Villegas et al. \cite{villegas17} combine both high-level future representation learning and pixel level prediction. These works are either focused purely on future video frame prediction or utilize future prediction as a self-supervised pre-training task. We, however, formulate our task as a multi-task learning problem. Further, future prediction solely at the pixel level \cite{ranzato2014, srivastava2015, lotter2017} have only been tested on short duration videos lasting a few seconds, while laparoscopic videos can have a duration of a few hours. Villegas et al. \cite{villegas17} show that predicting high-level structure within videos is beneficial for future prediction in long duration sequences. In this work, we predict the high-level visual representation of future frames of a video using LSTM networks. We believe that the ability of an LSTM to capture temporal information is beneficial for predicting how a sequence develops over time.

\subsection{Action Anticipation}
Our task also falls under the category of action anticipation as we aim to recognize the surgery before the entire surgery is complete. Action anticipation is quite an important task for applications of computer vision which require a system to act well in advance of the upcoming events. Typical examples would be Autonomous driving and surveillance systems. Previous works on this problem have focused on modifying loss functions so as to penalize the model more heavily as it makes mistakes at later times steps. Jain et al. \cite{jain2016} attached an exponentially increasing weight to the loss to achieve this so that the model is urged to make adjustments early into the video. More recently, Aliakbarian et al. \cite{aliakbarian2017iccv} argued that the penalty on false positives should be less during the initial time-steps of the video and increased over time to give equal weight to false positive and negatives at the last time step. The intuition behind this novel loss function is to encourage the model to make correct predictions early while accounting for possible ambiguities in the prediction at the early stages. Their loss function achieves superior accuracy to previous works on datasets like UCF-101 and JHMDB-21 with minimal video observation. We experiment with such modifications to standard loss functions to achieve our goal.

\section{Methodology}\label{sec:method}

\begin{figure}[t]
\centering
\includegraphics[scale=0.4]{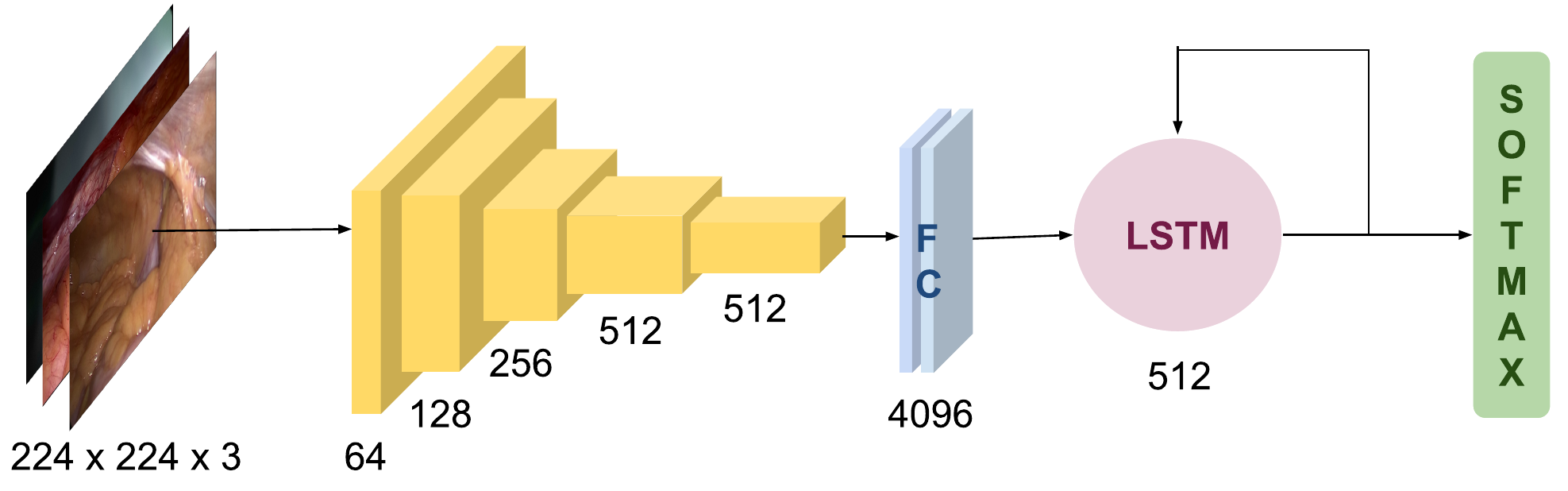}
\caption{High-level representation of our CNN+LSTM architecture.}
\label{fig:basic_model}
\end{figure}

\subsection{Model}

Our proposed model consists of a CNN+LSTM architecture shown in Fig. \ref{fig:basic_model}. While the CNN captures spatial information within the video frames, the LSTM captures temporal information related to the evolution of a surgery over time. The CNN+LSTM model is also well suited for the task of early recognition as predictions are obtained at every time-step. This work uses the VGG-16 version D from \cite{simonyan_vgg14}, pre-trained on the ImageNet dataset, as the CNN. We also explored two other networks, AlexNet and Resnet-50, and found VGG to perform the best.

As laparoscopic videos are of very long durations, the CNN+LSTM network cannot easily be trained end-to-end on complete video sequences due to memory restrictions. Hence, a two-step optimization approach is utilized where features for the video frames are first extracted from a fine-tuned CNN. These features are then directly provided to the LSTM, which is trained on complete video sequences. A VGG-16 network is initially fine-tuned on our task and then features of size 4096 are extracted from the penultimate layer in the VGG-16 architecture and fed to the LSTM. We use a single LSTM cell with a hidden-state size of 512 in our architecture which is followed by a fully-connected layer of size 9 (equal to the number of classes). The softmax function is applied at the output of the fully-connected layer.

With the aim of improving early recognition performance, we propose methods for improving the performance of both the CNN and the LSTM network. Our proposed approaches complement each other and result in the best early recognition performance.

\begin{figure*}[ht]
\centering
\includegraphics[scale=0.7]{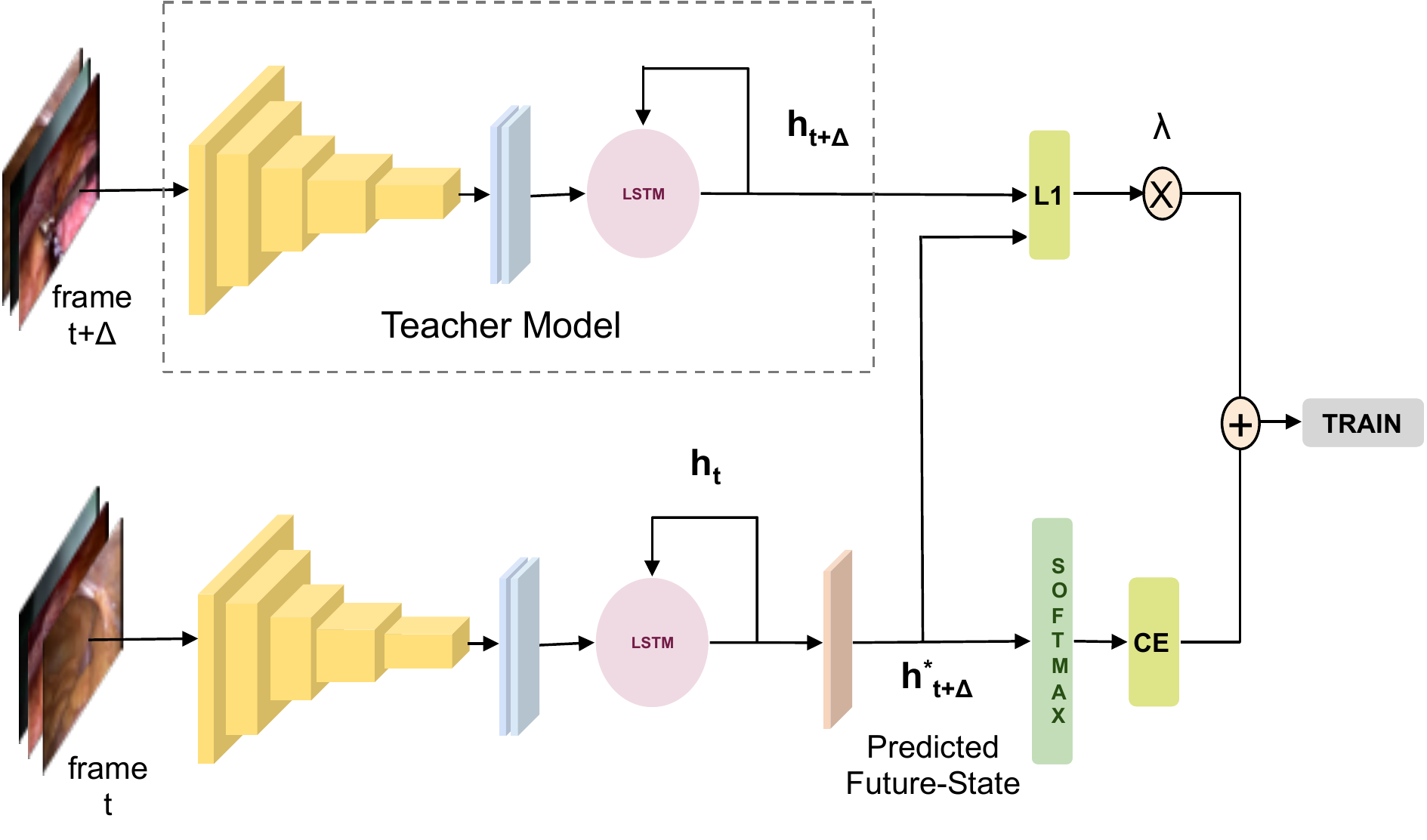}
\caption{Visualization of the {\ours} training scheme. In addition to surgery recognition, the FSP-LSTM predicts the future hidden state, namely the state at time $t+\Delta$, of the teacher model.}
\label{fig:state_pred_model}
\end{figure*}

\subsection{Proposed CNN Fine-Tuning}\label{method:finetune}

We fine-tune our CNN on the surgery recognition task with the aim to obtain more informative features, which can later be fed to the LSTM. For early recognition, it is beneficial for the CNN to capture discriminative features from frames that appear in the beginning of the laparoscopic video. Additionally, as discussed in section \ref{sec:related_work}, CNNs can be optimized for video classification tasks by incorporating data from across an entire video, rather than just utilizing single frames. To this end, we propose to train the CNN to extract features from complete videos while laying emphasis on the initial frames of the video.

Instead of using single frames of the video as input to the CNN, we stack a set of sampled frames spanning the entire video. Moreover, the sampling is performed for every iteration of training, thereby enabling the CNN to see different combinations of frames in the course of its fine-tuning and also serving as source of regularization. In order to sample frames, the video is first divided into segments of 200 frames each (corresponding to 200 seconds of the video since the dataset is sampled at 1fps). Then, 2 frames are uniformly sampled from each segment. This is essentially the same as sampling 1\% of the video except that we first divide the video into segments to ensure that frames are sampled from the entire length of the video. This sparse sampling strategy is adopted for two reasons. Firstly, our dataset consists of videos up to a length of 14000 frames and the GPU memory limits us from using a batch size of greater than 150 frames with the VGG-16 network, hence giving us the sampling rate of 1\%. Secondly, previous works \cite{wang2016, chen2017} have shown that sparse temporal sampling can be an effective and computationally efficient way of incorporating temporal information in the input to CNNs.

To obtain an aggregate feature vector for the sampled frames of a video, we insert a weighted max-pooling layer before the softmax classification layer in our CNN architecture, similar to \cite{ng2015}, so as to pool features over the temporal dimension of the input. The weights are set higher for the earlier frames as compare to the latter frames, in order to focus on early recognition. Formally, the weighting scheme is represented by Eq. \ref{eq:weighted_max_pooling}:

\begin{align}
\label{eq:weighted_max_pooling}
& F = \max_{t \in T_s} (w_t f_t),\\
& w_t = \begin{cases}
	              1, & 1 \leq t < \frac{T}{4}\\
				  0.5, & \frac{T}{4} \leq t < \frac{T}{2}\\
				  0.25, & \frac{T}{2} \leq t < \frac{3T}{4}\\
				  0.125, & \frac{3T}{4} \leq t \leq T
	              \end{cases} \nonumber
\end{align}
where $f_t$ is the feature vector output by the last fully-connected-4096 layer (post-activation) of VGG-16 for the $t^{th}$ frame of the video, $F$ is the feature vector after the weighted max-pooling over $f_t$'s across the time dimension. $T$ is the total number of frames in the video and $T_s$ is the set of all sampled frames from the video. $F$ is then fed to a softmax layer for classification. It is important to mention here that the sub-sampling and weighted max-pooling operations described above are done only during fine-tuning of the CNN. While training the LSTM, we make use of all the frames in the video and directly use $f_t$ as input to the LSTM.
\begin{figure*}[t]
\centering
\includegraphics[scale=0.5]{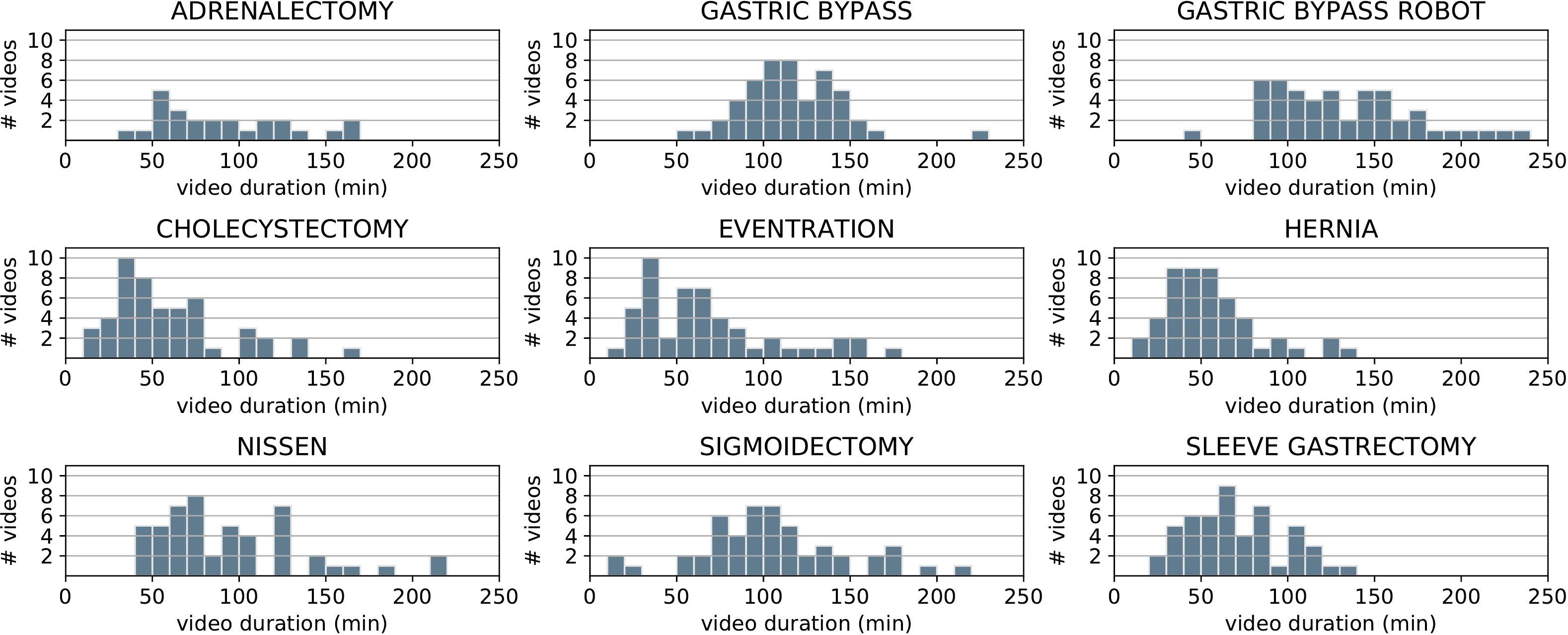}
\caption{Class-wise distribution of surgery video lengths}
\label{fig:dataset_class_wise_spread}
\end{figure*}

\begin{figure*}[ht]
	\begin{subfigure}{0.425\linewidth}
		\begin{subfigure}{0.32\linewidth}
			\centering
			\includegraphics[width=0.9\linewidth]{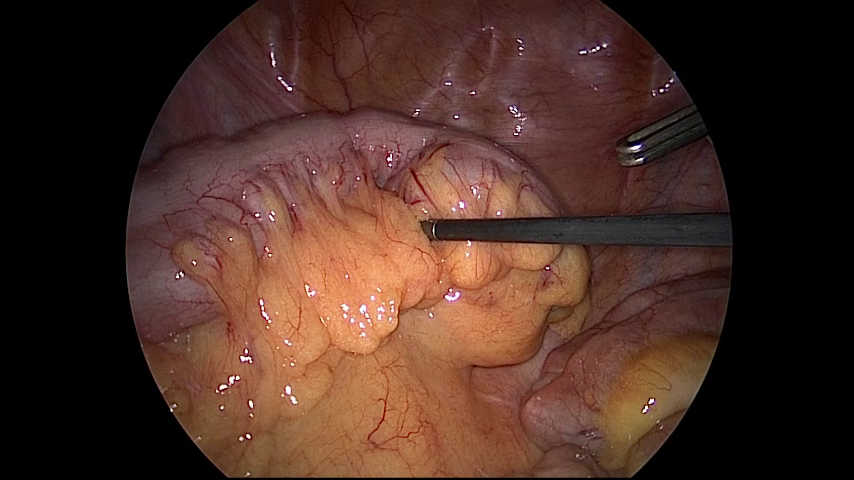}
		\end{subfigure}
		\begin{subfigure}{0.32\linewidth}
			\centering			\includegraphics[width=0.9\linewidth]{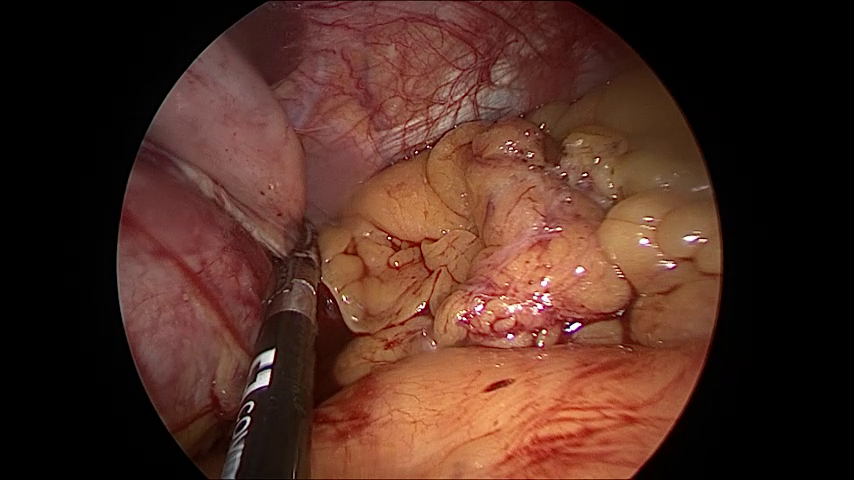}
		\end{subfigure}
		\begin{subfigure}{0.32\linewidth}
			\centering
			\includegraphics[width=0.9\linewidth]{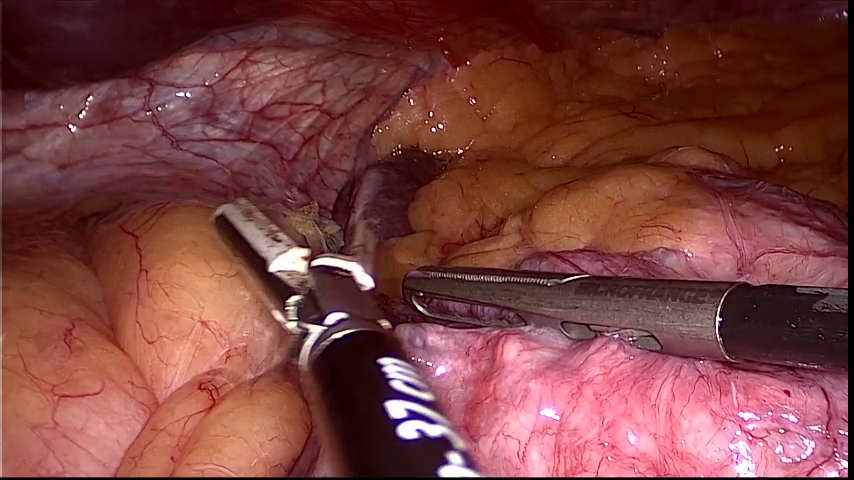}
		\end{subfigure}
	\caption{Similarity of frames within different surgery types}
	\label{fig:inter_class_similarity}
	\end{subfigure}
	\begin{subfigure}{0.575\linewidth}
		\begin{subfigure}{0.242\linewidth}
			\centering			\includegraphics[width=0.9\linewidth]{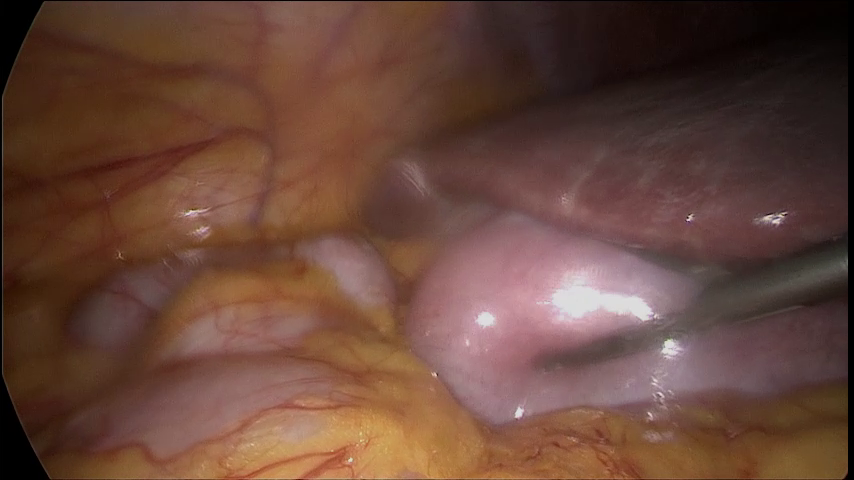}
			\caption{Motion Blur}
		\end{subfigure}
		\begin{subfigure}{0.242\linewidth}
			\centering			\includegraphics[width=0.9\linewidth]{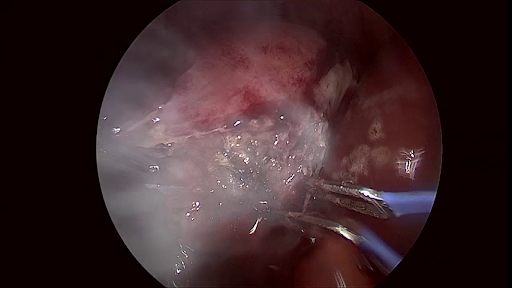}
			\caption{Smoke}
		\end{subfigure}
		\begin{subfigure}{0.242\linewidth}
			\centering			\includegraphics[width=0.9\linewidth]{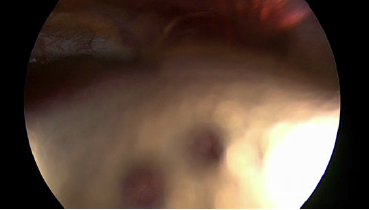}
			\caption{Blood stain}
		\end{subfigure}
		\begin{subfigure}{0.242\linewidth}
			\centering
\includegraphics[width=0.9\linewidth]{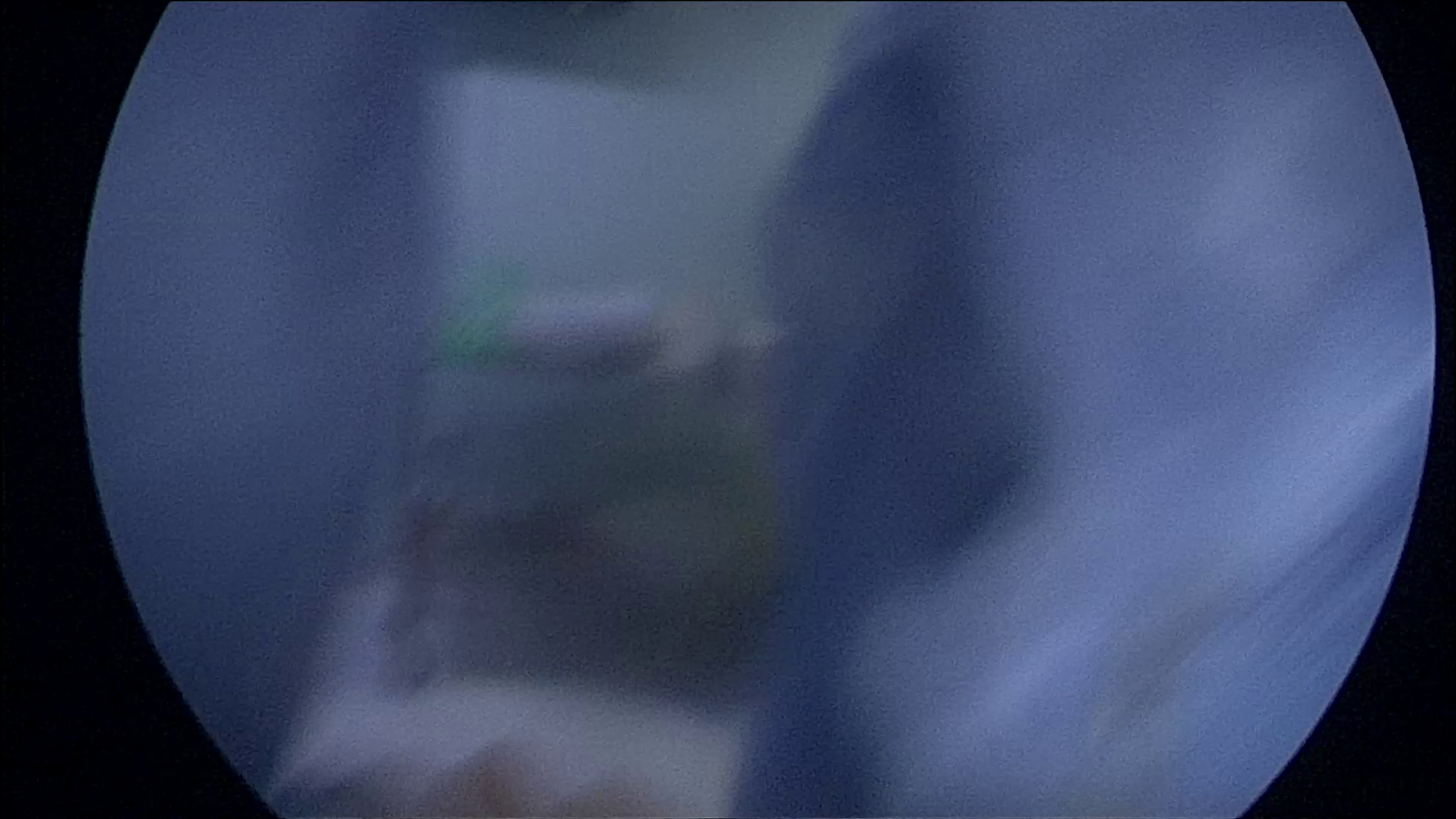}
			\caption{Irrelevant frame}
		\end{subfigure}
	\end{subfigure}
    \caption{Challenges associated with Laparo425 dataset}
    \label{fig:challenges}
\end{figure*}

\subsubsection*{Comparisons}
To highlight the performance improvement obtained by our proposed CNN fine-tuning scheme, we compare our proposed approach with the following methods for CNN fine-tuning:

\begin{enumerate}
\item \textbf{No fine-tuning:}
The LSTM is trained on features extracted from the video frames by a CNN pre-trained only on ImageNet. This can be considered as a basic approach, since the CNN is not fine-tuned on the task of surgery recognition. This comparison helps us to understand the advantage of fine-tuning the CNN on the surgery recognition task.

\item \textbf{Single-frame:}
In the second method, we fine-tune the CNN on individual frames of each video to predict the type of surgery they belong to. This can be considered as rather a na\"ive approach for fine-tuning the CNN given that the frames from different surgery types are very similar in appearance and it may be difficult for the CNN to discriminate between surgeries without any temporal knowledge of work-flow of the surgery. 

\item \textbf{Unweighted sub-video:}
We fine-tune the CNN for surgery recognition on sub-videos, similarly to our proposed approach, but without weighting scheme. 1\% of a laparoscopic video is extracted as described before and temporally discriminative information is obtained by performing a max-pooling operation, however without any weighting. We aim to study the benefit of weighted max-pooling for early recognition performance.

\end{enumerate}

\subsection{\ours}\label{sec:state_predicting_lstm}

Since our task is to recognize the type of surgery early, we explore if it is beneficial to train our model to predict events into the future. This can encourage the model to anticipate the evolution of the surgery and potentially obtain more discriminative information at the early stages of the surgery. This argument is further strengthened by the fact that Twinanda et al. \cite{twinanda2014} report higher classification accuracies when observing the later part of a laparoscopic surgery, the last 20\%, as compared to the initial part, the first 20\%.

\begin{table*}[t]
\small
\begin{center}
	\begin{tabular}{|c|c|c|c|c|c|c|c|c|c|}
    	\hline
		\multirow{2}{*}{Model}
        &\multirow{2}{*}{Epochs}
        &\multirow{2}{*}{Optimizer}
        &\multirow{2}{*}{Batch size}
		&\multicolumn{5}{c|}{Hyper-parameters}\\
        \cline{5-10}
		&&&& Learning rate & Momentum & Weight decay & Dropout\\
        \hline
        CNN fine-tuning on single frames & 30 & \multirow{6}{*}{SGD} & 128 & $10^{-3}$ & \multirow{6}{*}{0.9} & $10^{-3}$ & \multirow{3}{*}{$0.5$}\\
        \cline{1-2}\cline{4-5}\cline{7-7}
        CNN fine-tuning on unweighted sub-video & 105 & & \multirow{5}{*}{variable} & \multirow{3}{*}{$10^{-5}$} & & \multirow{3}{*}{$10^{-4}$} & \\
        \cline{1-1}\cline{2-2}
        CNN fine-tuning on weighted sub-video & 370 & & & & & & \\
        \cline{1-1}\cline{2-2}\cline{5-5}\cline{7-8}
        LSTM training & 145 & & & \multirow{3}{*}{$10^{-3}$} & & \multirow{3}{*}{$10^{-3}$} & \multirow{3}{*}{$0.0$} \\
        \cline{1-1} \cline{2-2}
        FSP-LSTM training, $\Delta$=0.2, $\lambda$=10 & 105 & & & & & &\\
        \cline{1-1} \cline{2-2}
        FSP-LSTM training, $\Delta$=10 min, $\lambda$=100 & 75 & & & & & & \\
        \hline
	\end{tabular}
\end{center}
\caption{List of hyper-parameter values (FSP=Future-State Predicting)}\label{tab:hparams}
\end{table*}
\medskip
\noindent\textbf{Model architecture:} We introduce a novel framework to incorporate this idea. We train an LSTM in a multi-task manner to predict future visual representations along with the surgery type. The output of the LSTM is provided to a fully-connected layer, which is used to replicate the visual representation of a future frame. The future visual representation is obtained from the hidden state of a teacher LSTM model that has been previously trained on the surgery recognition task alone. The features from the proposed fine-tuned CNN is provided as input to both the {\ours} and the teacher LSTM model. Fig. \ref{fig:state_pred_model} illustrates the proposed framework. Here, $h_t$ is the hidden state output by the {\ours} (FSP-LSTM), $h_{t+\Delta}$ is the true hidden state of the teacher model at time-step $(t+\Delta)$ and $h^{*}_{t+\Delta}$ is the predicted hidden state at time-step $(t+\Delta)$, which is to be learned by the {\ours} at time-step $t$. A weighted combination of cross-entropy and regression loss is then used to train the student LSTM while the teacher LSTM is only used to provide ground-truth hidden states for the student LSTM to learn. We hypothesize that by learning to predict the hidden state at a future time-step our model can now attain an accuracy close to that obtained by our previous models, but earlier.

\medskip
\noindent\textbf{Choice of $\Delta$:} In this work we explore two variants of $\Delta$. First, we experiment with a variable $\Delta$ by setting it as a fraction of the total number of frames in the video. For instance, if $\Delta$ is 0.5, the model learns to predict the hidden state of the ($t+\Delta T$)th frame after observing the first $t$ frames, where $T$ is the total number of frames in the video. As a consequence of learning the future hidden states, when we reach the $(1-\Delta)T$th frame, we switch over to training on the next video as the remaining frames for the current video do not have a future hidden state to learn. This is acceptable to us since the task we are interested in is the early recognition of the surgery type. Hence, we assume that our model does not wait to see the end-frames of the surgery, rather, it seeks to predict information from these frames before their actual occurrence. In the second variant, $\Delta$ is measured in minutes and remains constant for all videos. For instance, if $\Delta$ is 5 minutes, then the model learns to predict the hidden state at the $(t+ 60\Delta)$th frame, as the frames are sampled at 1fps, after observing the first $t$ frames. Similarly as before, when the $(T- 60\Delta)$th frame is reached, the training is switched to the next video.

\medskip
\noindent\textbf{State-prediction loss:} We formulate the multi-task loss function as follows:

\begin{equation}
\label{eq:state_pred_loss}
L(y, \hat{y}) = L_{Classification} + \lambda L_{Future Prediction},
\end{equation}
where $L_{Classification}$ is the loss function described in section \ref{subsec:early_loss} below and for $L_{FuturePrediction}$, which is the error in prediction of the future visual representation, we experiment with two different loss functions, namely smooth L1 and L2. $\lambda$ is a hyper-parameter that needs to be tuned to yield a loss function that optimally weights the two tasks. Note that the teacher LSTM model is picked from previous experiments, but not trained while performing this experiment. Only the weights of the student model are optimized.

\subsection{Loss function for early recognition}
\label{subsec:early_loss}
We utilize the cross-entropy (CE) loss function, Eq. \ref{eq:avg_loss}, for the surgery recognition task, computed at each time-step and averaged before back-propagating through the entire video. The {\it average loss} function is given as:

\begin{equation}
\label{eq:avg_loss}
L(y, \hat{y}) = \frac{1}{T}\sum_{k=1}^{N}\sum_{t=1}^{T}y^k_t\log{\hat{y}^k_t},\\
\end{equation}

\noindent where $T$ is the total number of frames for a given video and $N$ is the number of classes, $y^k_t$ is the ground truth one-hot encoded label value for class $k$ at time-step $t$ and $\hat{y}^k_t$ is the corresponding prediction confidence of the model.

To explore the advantage of the previous approaches that modify the standard py loss function for improving early recognition performance, we also use the {\it linear weighted average loss} function proposed by Aliakbarian et al. \cite{aliakbarian2017iccv}:

\begin{equation}
\label{eq:lin_wt_avg_loss}
L(y, \hat{y}) = \frac{1}{T}\sum_{k=1}^{N}\sum_{t=1}^{T}y^k_t\log{\hat{y}^k_t} + \frac{t(1 - y^k_t)}{T}\log{(1 - \hat{y}^k_t)}\\
\end{equation}

The high-level idea behind this loss function is to reduce the penalization on the false positives for the initial time-steps and progressively increase the penalization as the model sees more of the video. The justification provided by the authors in \cite{aliakbarian2017iccv} for this proposed scheme is that actions like running and high-jump may be ambiguous at the starting and can be easily confused. Hence, predicting high-jump instead of running at the start does not call for a heavy penalization of weights. At the same time, it is also desired that the correct class be predicted early with high probability and therefore, penalization on false negatives is not reduced. The same argument can be extended to our dataset as well. The initial procedures in the different laparoscopic surgeries are similar because they involve the introduction of the trocars into the abdomen to make the internal anatomical structures accessible by surgical tools.

\section{Experimental setup} \label{sec:exp_setup}
\subsection{Laparo425 Dataset}
Our dataset consists of 425 videos belonging to 9 different classes of laparoscopic surgeries performed at the University Hospital of Strasbourg/IHU. The videos were down-sampled to 1 fps for our experiments. We split our dataset into three sets for training, validation and testing in the ratio of 60\%(254), 10\%(43) and 30\%(128) respectively. Table~\ref{tab:dataset} summarizes the number and average duration of the videos for each surgery class in the dataset. Fig.~\ref{fig:dataset_class_wise_spread} illustrates the class-wise distribution of the video durations for different surgeries.

\begin{table}[H]
\small
\begin{center}
\begin{tabular}{|c|c|c|}
\hline
\multirow{1}{*}{Surgery Type} & \multirow{1}{*}{\# Videos} & \multirow{1}{*}{Duration (min.)} \tabularnewline
\hline
\hline
Adrenalectomy & 25 & 86.4 $\pm$ 36.4 \tabularnewline
Gastric Bypass & 50 & 113.8 $\pm$ 29.3 \tabularnewline
Gastric Bypass Robot & 50 & 129.5 $\pm$ 40.0 \tabularnewline
Cholecystectomy & 50 & 57.7 $\pm$ 32.3 \tabularnewline
Eventration & 50 & 67.5 $\pm$ 39.0 \tabularnewline
Hernia & 50 & 54.8 $\pm$ 25.5 \tabularnewline
Nissen & 50 & 92.2 $\pm$ 42.2 \tabularnewline
Sigmoidectomy & 50 & 102.8 $\pm$ 41.8 \tabularnewline
Sleeve Gastrectomy & 50 & 69.4 $\pm$ 26.3 \tabularnewline
\hline
\textbf{Total} & 425 & 86.0 $\pm$ 43.4 \tabularnewline
\hline
\end{tabular}
\end{center}
\caption{List of surgery types, number of videos present in each type and mean duration of the videos per type ($\pm$ std)}
\label{tab:dataset}
\end{table}

\subsection{Training approach}

\medskip
\noindent\textbf{Data pre-processing and augmentation:}
The images in our dataset are of size 256x256. For training the CNN, random crops of size 224x224 are performed on the images and Imagenet mean RGB values are subtracted from the respective channels of the image. And for LSTM training, the CNN features are extracted using 224x224 centre crops of the original images. The same procedure is followed while testing the LSTM as well.

\begin{table*}[t]
	\begin{center}
    	\begin{tabular}{|c|c|c|c|c|c|c|c|c|}
        \hline
        \multicolumn{2}{|c|}{Model}
       &\multicolumn{7}{c|}{Accuracy (\%)}\\
        \hline
        CNN features & Loss & 2 min & 4 min & 6 min & 8 min & 10 min & 12 min & 14 min\\
        \hline
        No Fine-tuning & Avg. & 42.97 & 59.38 & 57.81 & 61.72 & 64.06 & 63.28 & 64.84 \\ \hline
        Single frames & Avg. & 46.09 & 50.78 & 61.72 & 60.16 & 62.50 & 60.16 & 62.50 \\ \hline
        Unweighted Sub-video & Avg. & 37.50 & 46.09 & 53.12 & 61.72 & 66.41 & 68.75 & 71.10 \\ \hline
		Weighted sub-video & Avg. & \textbf{53.91} & 57.03 & \textbf{65.62} & 66.41 & \textbf{71.09} & \textbf{73.44} & 74.22 \\
 \hline
		Weighted sub-video & Lin. Wt. Avg. & 51.56 & \textbf{59.38} & 60.16 & \textbf{67.97} & 69.53 & 72.66 & \textbf{75.78} \\ \hline
        \end{tabular}
	\end{center}
    \caption{Early recognition performance of LSTM trained with different CNN features and loss functions but without state-prediction.}
\label{tab:results_without_state_prediciton}
\end{table*}

\begin{table*}[t]
	\begin{center}
		\begin{tabular}{|c|c|c|c|c|c|c|c|c|c|}
          \hline
          \multicolumn{3}{|c|}{Model}
         &\multicolumn{7}{c|}{Accuracy}\\
         \hline
         LSTM & $\Delta$ & $\lambda$ & 2 min & 4 min & 6 min & 8 min & 10 min & 12 min & 14 min\\
         \hline
         Teacher LSTM & - & - & \textbf{51.56} & \textbf{59.38} & 60.16 & 67.97 & 69.53 & 72.66 & 75.78 \\ \hline
         FSP-LSTM & 0.2 & 10 & 42.19 & 57.81 & \textbf{66.41} & \textbf{70.31} & \textbf{75.00} & \textbf{77.34} & 78.91 \\ \hline
         FSP-LSTM & 10 min & 100 & 44.53 & 56.25 & \textbf{66.41} & 67.97 & 73.44 & 75.78 & \textbf{80.47}\\ \hline
        \end{tabular}
	\end{center}
    \caption{Early recognition performance using state-prediction (FSP=Future-State Predicting)}
\label{tab:results_with_state_prediciton}
\end{table*}

\begin{figure*}[!t]
  \begin{subfigure}{0.49\linewidth}
  \includegraphics[scale=0.27]{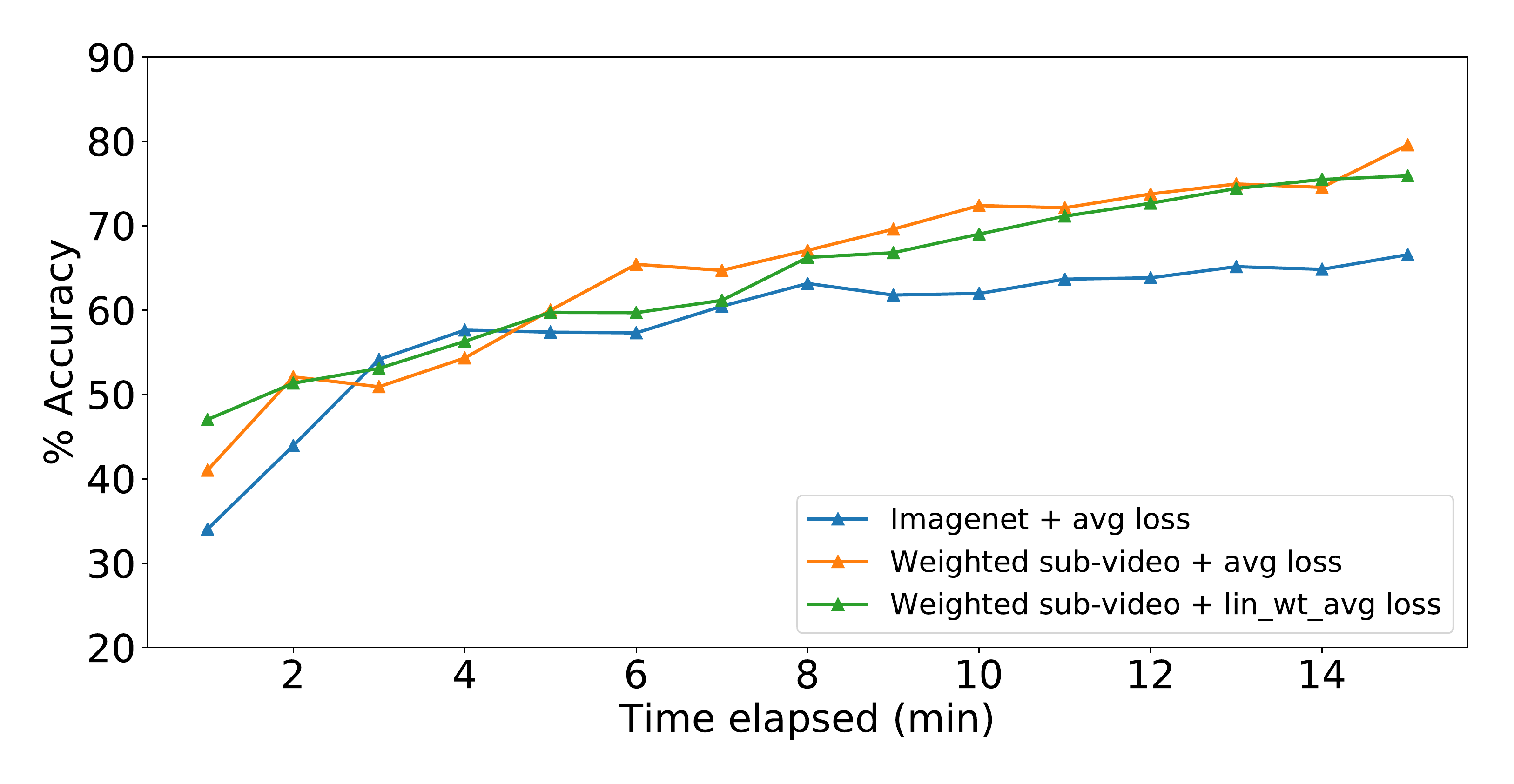}
  \caption{Early recognition performance without state prediction}
  \label{fig:plot_early_rec_wo_state_pred}
  \end{subfigure}
  \begin{subfigure}{0.49\linewidth}
  \includegraphics[scale=0.27]{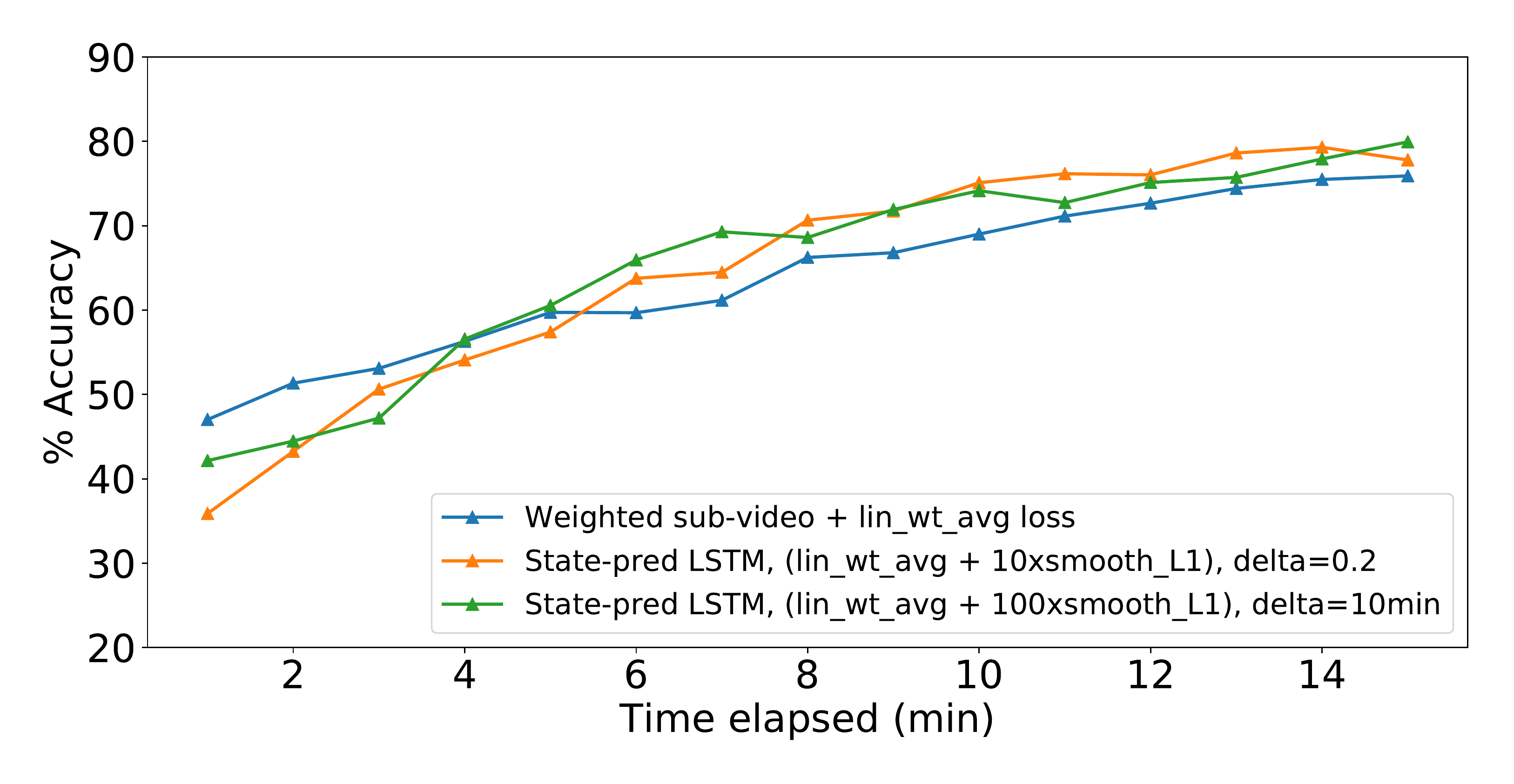}
  \caption{Early recognition performance with state prediction}
  \label{fig:plot_early_rec_w_state_pred}
  \end{subfigure}
  \caption{Plot of classification accuracy vs time elapsed in the surgery}
\end{figure*}

\medskip
\noindent\textbf{Hyper-parameter search:}
In order to obtain the optimal hyper-parameters for training our model, an extensive grid search over the hyper-parameter search space was performed. We varied the values of learning rate, weight decay, dropout ratio, LSTM state size, $\lambda$ from Eq. \ref{eq:state_pred_loss} and $\Delta$ for the \ours. The optimal values so obtained are listed in Table \ref{tab:hparams}. Variable batch sizes have been used for most of the experiments because we train our models on variable length videos. In the case of CNN fine-tuning on sub-videos, we sample 1\% of the frames in a given video during each step of training. For LSTM training, the CNN features of all the frames of a given video are fed to the LSTM for a training step. Consequently, the number of frames in a batch varies with each video. The parameters are summarized in Table \ref{tab:hparams}.

\subsection{Evaluation metric}
We assess the performance of our model by evaluating the recognition accuracy against the elapsed time of the surgery. This demonstrates the performance of a model from a practical standpoint. Note that while plotting elapsed time on the x-axis we are limited to the length of the shortest video for reporting average accuracies, i.e. 15 minutes. 

\section{Results \label{sec:results}}

\subsection{Early Recognition CNN}
We present the classification accuracies of the various approaches we adopt to improve the CNN features in Table \ref{tab:results_without_state_prediciton}.
As expected, the models perform increasingly better as they see more of the video. But we observe that the trend is not monotonically increasing, though the deviation is only minimal. This may be due to the presence of frames irrelevant to the surgery, which the LSTM tries to associate to a surgery type and thereby, corrupting the information accumulated in the cell state thus far. We can also observe from the table that training the CNN on sub-sampled frames of a video is critical to performance of the LSTM. Individual images in our dataset can be extremely difficult to classify for a CNN, or even an expert human for that matter, as they are visually very similar owing to the fact that they all belong to the domain of laparoscopic surgeries, which are performed in the abdominal area. However, the unweighted sub-video approach seems to perform relatively worse up until the first 6 minutes. We think this might be because of the max-pooling operation performed while fine-tuning the CNN. This may result in the features of the initial frames being suppressed since they may not be as discriminative as the features of the latter frames. Whereas, the single-frames approach uses each frame of the video for fine-tuning and hence, does not ignore the initial frames. Similarly, the weighted sub-video approach assigns greater weight to initial frames as discussed in section \ref{method:finetune}. Nevertheless, using unweighted sub-video features for the LSTM improves the accuracy by nearly 10\%-points over using single-frame features for the LSTM at the 14 minute mark. This suggests the importance of aggregating temporal information while fine-tuning the CNN. Furthermore, we see that weighting the features from earlier frames more heavily has indeed the desirable effect of improving early recognition performance of the LSTM. We achieve an improvement of  11\%-points over the model which is not fine-tuned, having observed only 2 minutes of the video, which translates to improvements at later time-steps as well. The margin of improvement is illustrated in Fig. \ref{fig:plot_early_rec_wo_state_pred}.

\subsection{Early recognition loss function}
We also assess the benefits of modifying the loss function to improve early recognition performance. In \ref{fig:plot_early_rec_wo_state_pred} we observe that the plots for the average loss function and for the linear weighted loss, Eq. \ref{eq:lin_wt_avg_loss}, yield similar accuracies for the most part. In fact, the average loss function even beats the linear weighted average loss by a notable margin near the 6 minute mark.

We reason out these outcomes of the proposed modifications to the loss function as follows: The long duration of our videos causes the loss function to reduce the penalization on the false positives for initial frames to a much greater degree than desired, leading to a counterproductive effect. To put this in perspective, an average video in our dataset consists of around 5000 frames (T in Eq. \ref{eq:lin_wt_avg_loss}). This means that the penalization on the first few frames is of the order of $10^{-4}$ (1/5000) for the linear weighted loss. On the other hand, the video clips in \cite{aliakbarian2017iccv} are typically a few seconds long and therefore, the initial penalization is only reasonably reduced. Nevertheless, we notice that the linear weighted average loss does perform better up until the first 4 minutes and manages to match the average loss, if not beat it, for most of the latter part. Hence, we choose the linear weighted loss for the \ours, which further improves early prediction performance.

\subsection{\ours}
Table \ref{tab:results_with_state_prediciton} and Fig. \ref{fig:plot_early_rec_w_state_pred} show the comparative performance of the FSP-LSTM with the teacher LSTM, trained on weighted sub-video features with the linear weighted loss. As the teacher LSTM has not been trained to predict the future, this experiment shows how training a model to predict future states improves early recognition performance. We observe that the FSP-LSTM, for both variants of $\Delta$, catches up to the teacher LSTM within the first 6 minutes and well surpasses it beyond this point. At the 14 minute mark, we are better off than the teacher LSTM by nearly 5\%-points and beat the basic approach, which does not fine-tune the CNN and neither incorporates future prediction, by a significant margin of 15\%-points using $\Delta$ as 10 minutes. \\

We find it intuitive that the FSP-LSTM lags the teacher LSTM initially because of the fact that our model has seen a very small fraction of the video based on which future states are to be predicted. Given that laparoscopic surgeries start off with the placement of the trocar into the abdominal cavity, it is hard to accurately predict the upcoming events in the surgery. But once we go past this phase, there arise variations in the tools used, region of operation and so forth, which enable the LSTM to better understand the context of the surgery and make a more accurate future state prediction.

\section{Discussions} 
\label{sec:discussion}

\subsection{Ablation Study}

We present an ablation study for the FSP-LSTM model to highlight the effect of the Delta and Lambda parameters as well as the loss function. In addition to this, the importance of training the model on the entire video is evaluated by comparing its performance to that of models trained on shorter durations of the video.

\subsubsection{Variation of Delta}

\begin{figure}[t]
    \centering
    \includegraphics[scale=0.275]{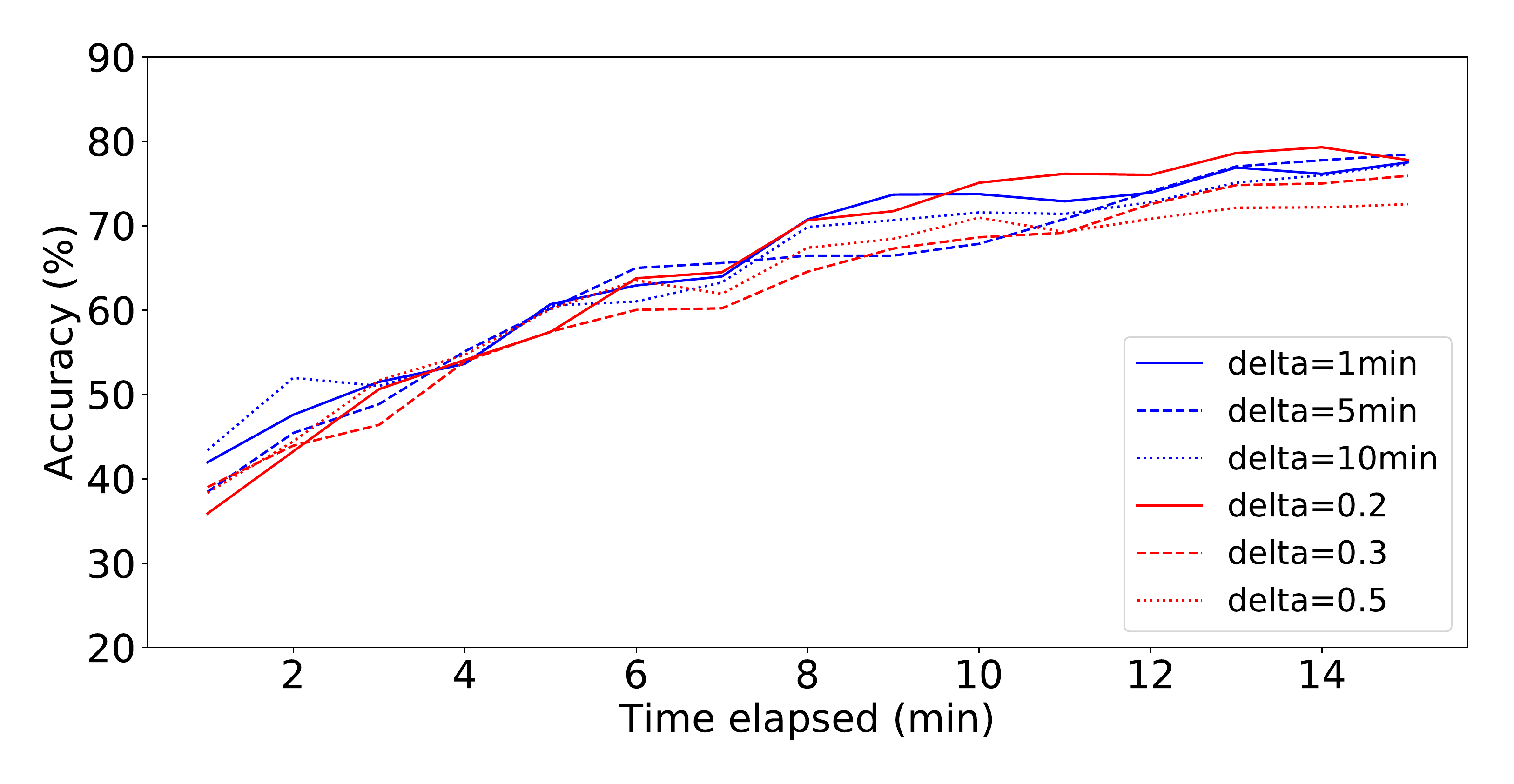}
    \caption{Early recognition performance of FSP-LSTM with varying $\Delta$. The smooth L1 loss is used and $\lambda=10$.}
    \label{fig:variation_of_delta}
\end{figure}

Different values of the $\Delta$ parameter are compared in Fig. \ref{fig:variation_of_delta}. Empirically we found the best $\Delta$ to be 10 min, when using a constant time $\Delta$, and 0.2 when $\Delta$ is considered as a fraction of the video length.

\subsubsection{Variation of Lambda}

\begin{figure}[t]
    \centering
    \includegraphics[scale=0.275]{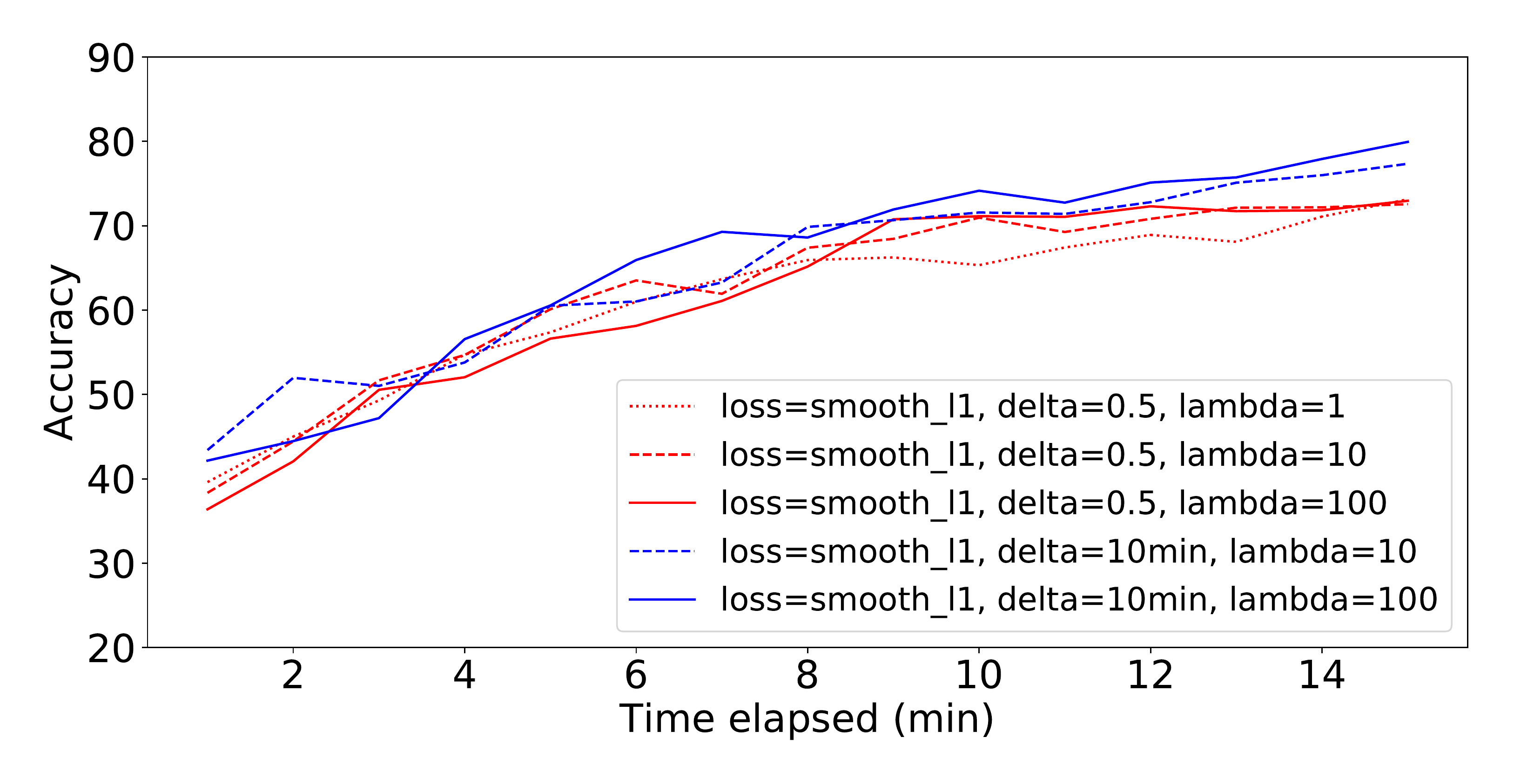}
    \caption{Early recognition performance of FSP-LSTM with varying $\lambda$}
    \label{fig:fig:variation_of_lambda}
\end{figure}

Fig. \ref{fig:fig:variation_of_lambda} illustrates the comparative performance of different settings of $\lambda$. The general trend we observed in our experiments was that when we increase the value of $\lambda$ the model is able to make better predictions of the future state, as is expected. But there is a trade-off in the sense higher the value of $\lambda$, later is the improvement, which is conflicting with our goal. If we were to reduce the value of $\lambda$, we attain small improvements early in the surgery but as time progresses, we fall short of the accuracy produced by a higher setting of $\lambda$. This observation makes sense because a high value of $\lambda$ essentially enforces a higher emphasis on the accuracy of future state prediction. It is relatively harder for the model to predict the future state accurately very early into the surgery as it has very little information to go with. But as time progresses, it gains more information and is able to better predict the future state, which in turn improves its classification performance. We see in Fig. \ref{fig:fig:variation_of_lambda} that this trend is manifested. When $\Delta=0.5$, the plot corresponding to $\lambda=100$ catches up with $\lambda=10$ only after 8min. Interestingly, when $\Delta=10$ min, the plot corresponding to $\lambda=100$ catches up with $\lambda=10$ after just 4min. This suggests that when $\Delta$ is lower, the $\lambda=100$ plot can surpass the $\lambda=10$ plot at an earlier point.

\subsubsection{Smooth L1 vs L2 Loss}
\begin{figure}[t]
    \centering
    \includegraphics[scale=0.275]{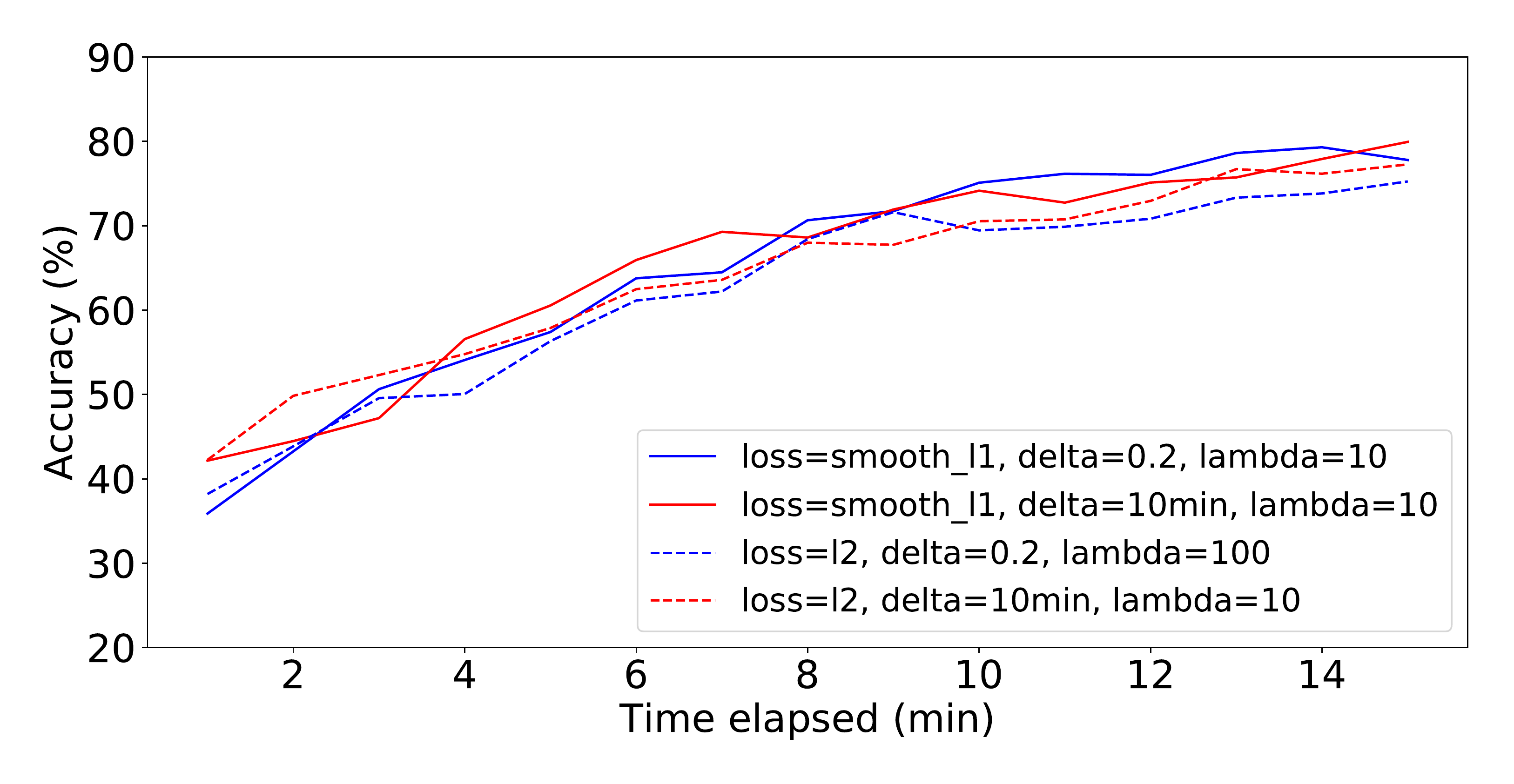}
    \caption{Early recognition performance of FSP-LSTM with smooth L1 and L2 losses}
    \label{fig:smooth_l1_vs_l2}
\end{figure}

Fig. \ref{fig:smooth_l1_vs_l2} illustrates the comparative performance of smooth L1 and L2 losses defining $L_{FuturePrediction}$. We find that the smooth L1 loss is better to train our model with as it gives us the benefits of both L1 and L2 losses. When the magnitude of error is less that 1, the smooth L1 mimics the L2 loss and the L1 loss otherwise. This prevents our model from being influenced too much by outliers. We found that when we have a large value of $\Delta$, a partial L1 behavior helps more than when $\Delta$ is smaller. This could be because the errors may tend to be higher as the model tries to learn the features of a more advanced time step in the surgery.

\begin{figure}[t]
    \centering
    \includegraphics[scale=0.275]{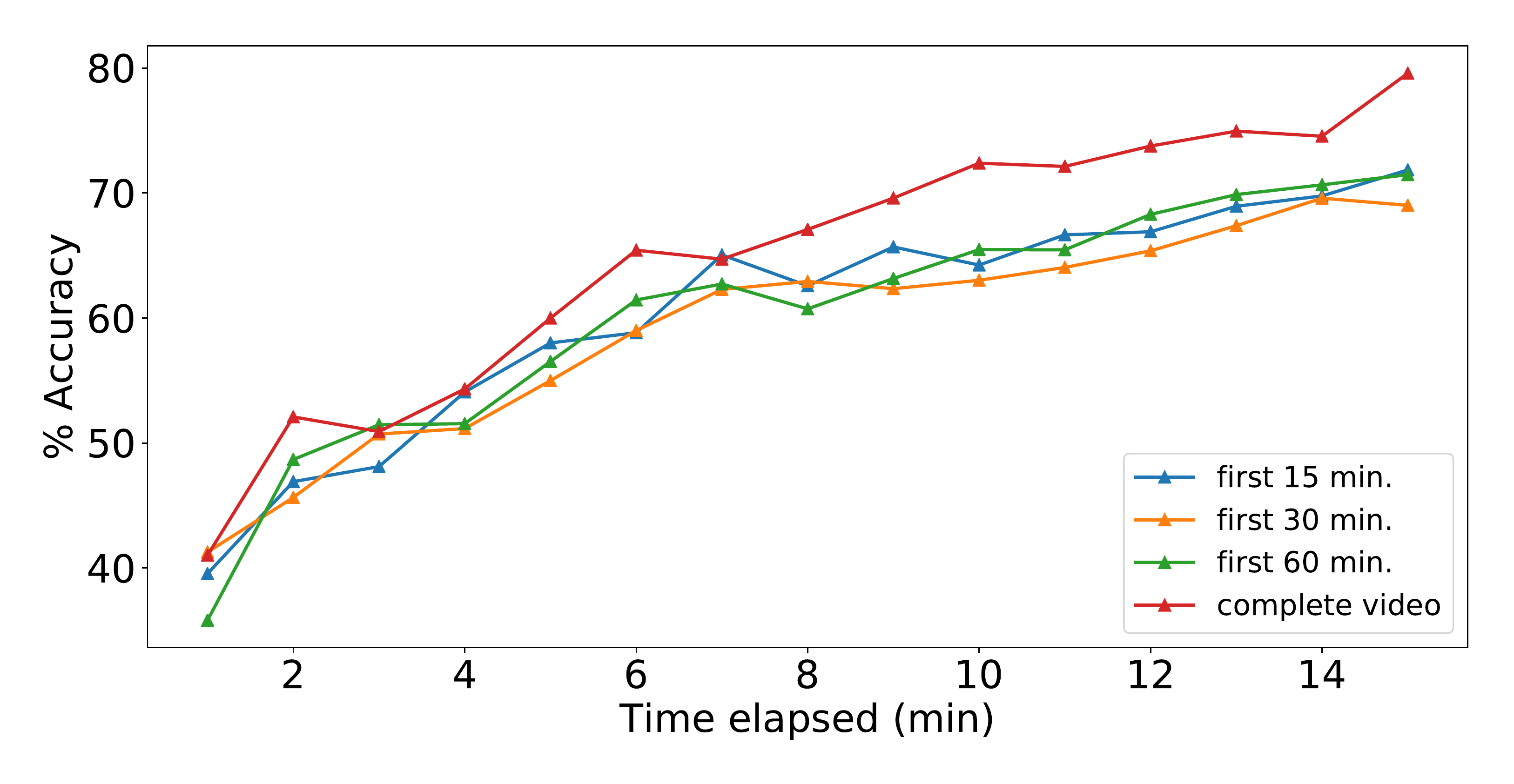}
    \caption{Comparative performance of LSTM + weighted sub-video variants:- training on first 15 min., 30 min., 60 min. and complete video}
    \label{fig:revision_result}
\end{figure}

\subsubsection{Importance of using complete videos for training}
The proposed CNN fine-tuning as described in \ref{method:finetune} sub-samples frames from the entire length of the video. Since our task is early recognition, we compare our fine-tuning method to approaches wherein the CNN is fine-tuned on just the first few minutes of the video. To this end, we compare 3 variants of the weighted sub-video scheme: fine-tuning on only the first 15 min., 30 min. and 60 min., to training on the complete video. To make up for the loss in number of frames sampled by maintaining a 1\% sampling rate, we increase the sampling rate to 10\%, 5\% and 2.5\% for 15 min., 30 min. and 60 min., respectively. The rest is exactly the same as the weighted sub-video CNN fine-tuning on the complete video. One other modification required is that at the time of LSTM training, we reduce the length of the input video sequence to the respective duration for which the CNN fine-tuning was performed.

The performance of the weighted sub-video variants trained on different video durations is presented in Fig. \ref{fig:revision_result}. The experiment clearly shows that the weighted sub-video scheme on the entire video significantly improves performance compared to the others despite increasing the sampling rate for the other variants. For this reason, we chose to carry out our experiments on future-state prediction using the complete video version. We suspect that forcing the model to extract discriminative features from the initial parts may have an undesirable effect on its generalization capabilities. But training on the complete video gives it the flexibility to ignore some information from the initial parts. This reasoning is supported by previous work \cite{twinanda2014} which has shown that the most discriminative parts of a laparoscopic surgery lie towards the last 20\% of the video. So we infer that it is better to have soft constraints like our weighting scheme described in \ref{method:finetune} instead of imposing a hard constraint that the model only pick up information from the initial parts of the video.

\begin{figure*}[t]
\centering
	\begin{subfigure}{\linewidth}
	\centering
		\begin{subfigure}{0.30\linewidth}
			\centering
			\includegraphics[width=\linewidth]{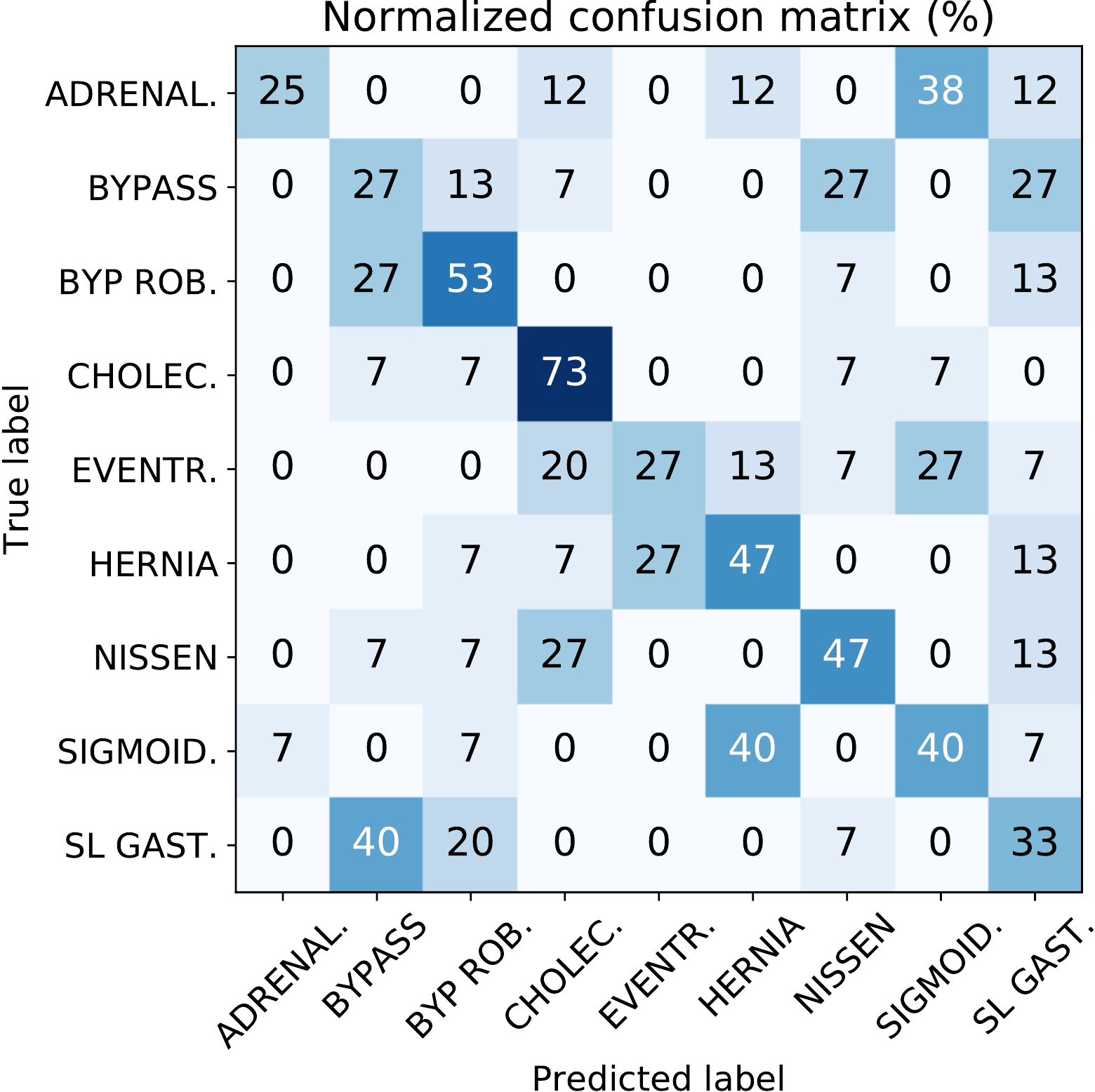}
		\end{subfigure}%
		\begin{subfigure}{0.30\linewidth}
			\centering			
			\includegraphics[width=\linewidth]{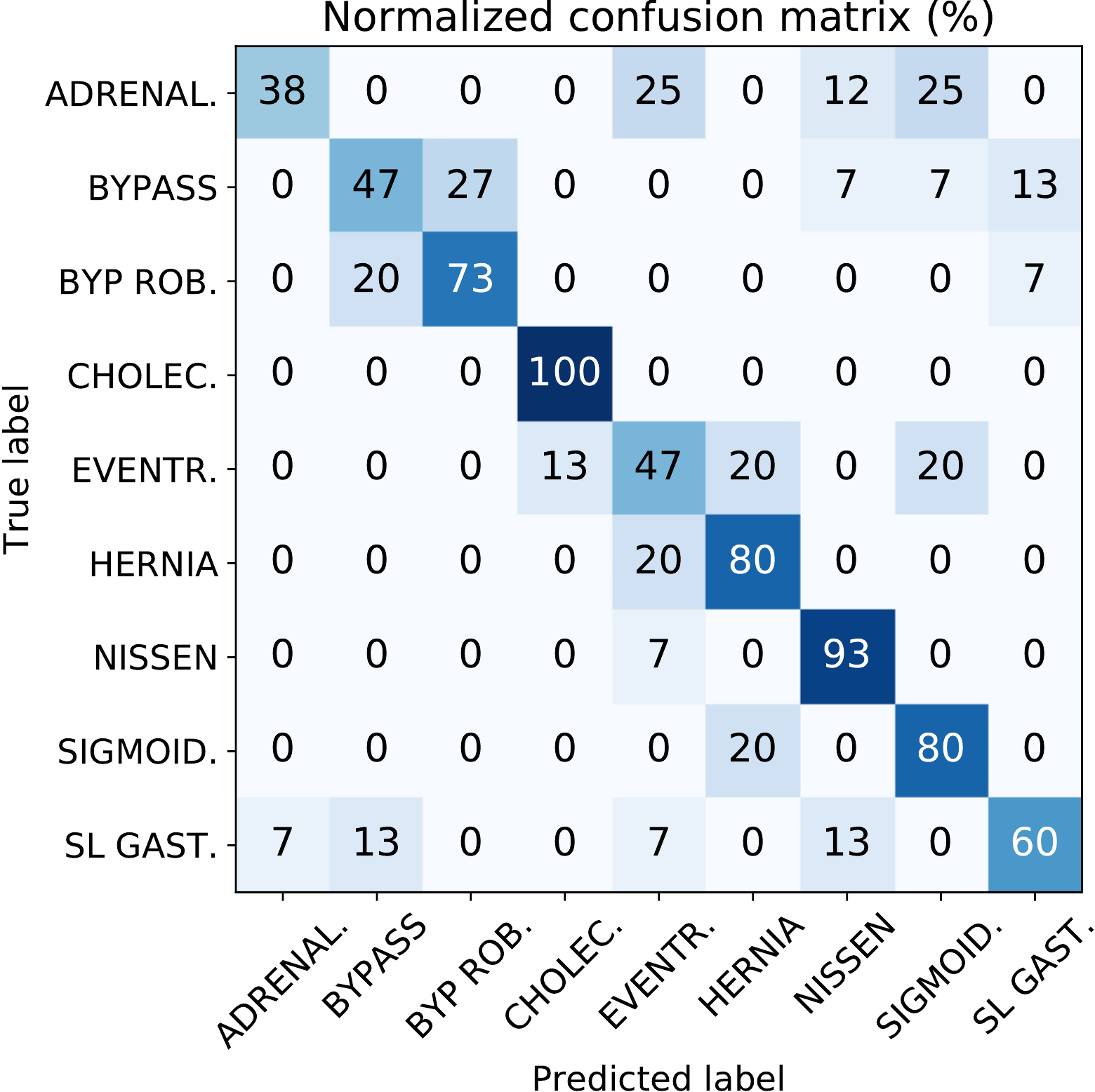}
		\end{subfigure}%
		\begin{subfigure}{0.30\linewidth}
			\centering
			\includegraphics[width=\linewidth]{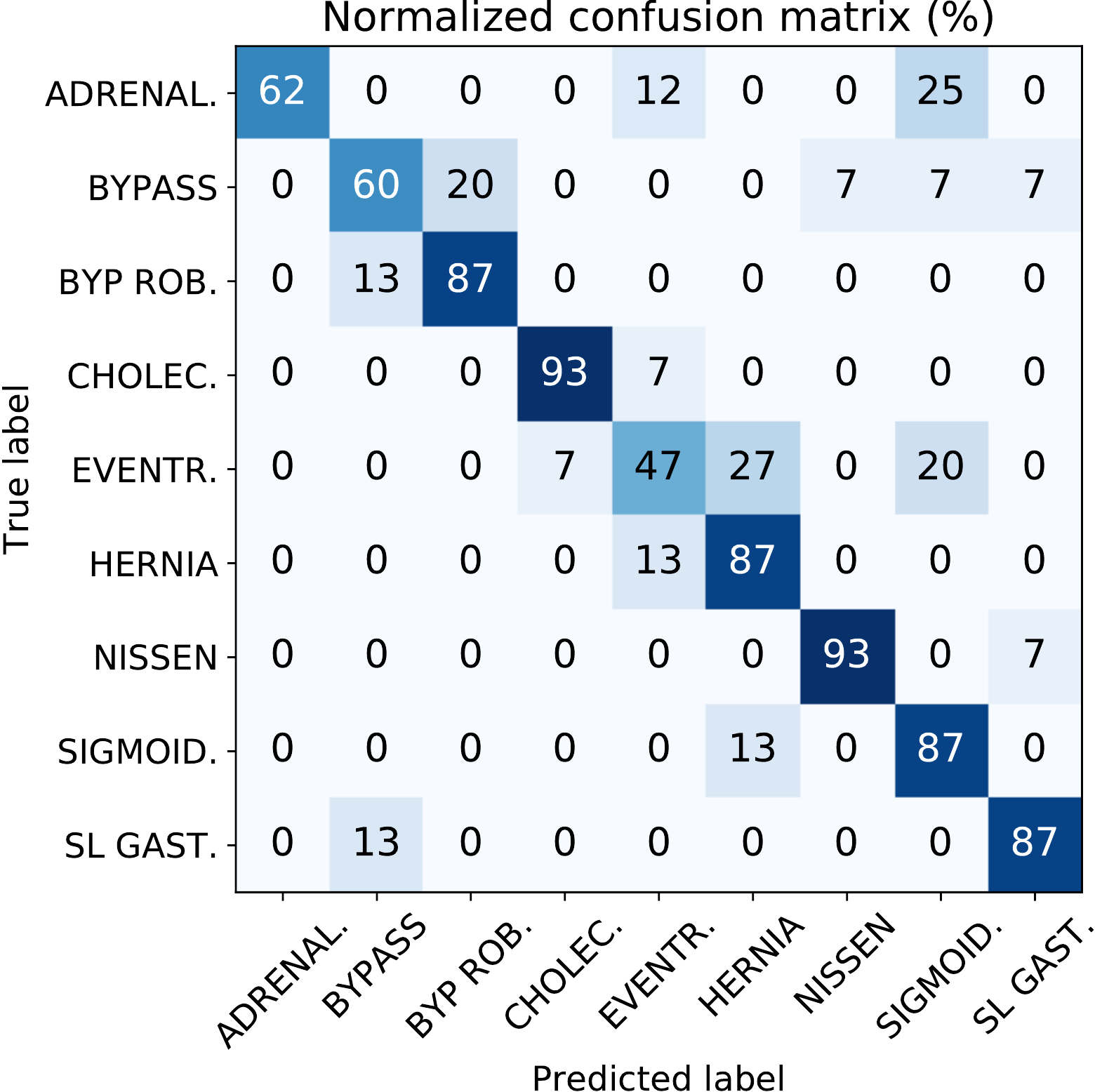}
		\end{subfigure}%
	\caption{Class-wise accuracy (\%) of {FSP-LSTM}  with $\Delta$=0.2 and $\lambda$=10 after 2min, 8min and 14min (from left to right)}
	\end{subfigure}
	
	\vspace{0.2cm}
	
	\begin{subfigure}{\linewidth}
	\centering
		\begin{subfigure}{0.30\linewidth}
			\centering
			\includegraphics[width=\linewidth]{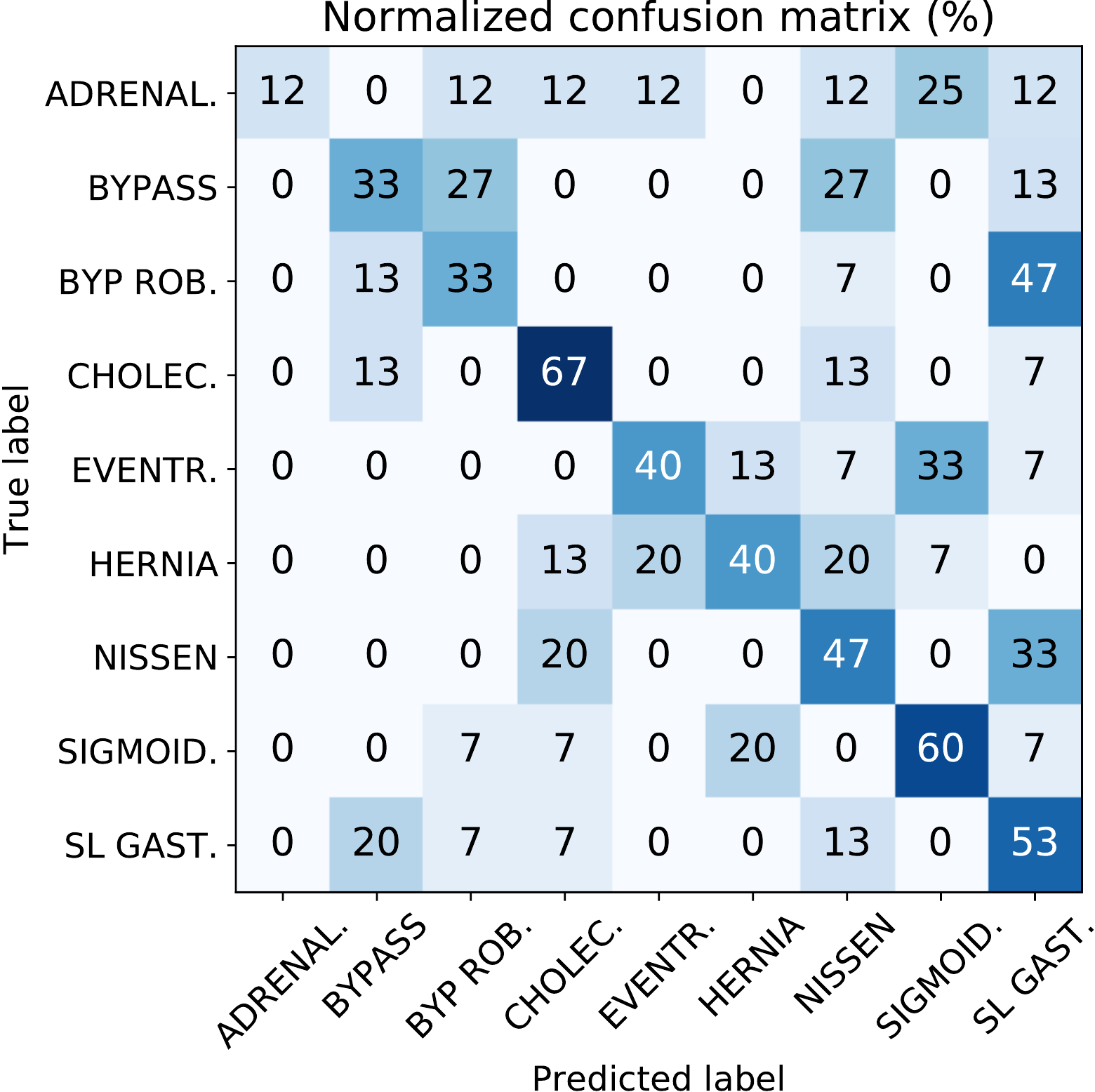}
		\end{subfigure}%
		\begin{subfigure}{0.30\linewidth}
			\centering			
			\includegraphics[width=\linewidth]{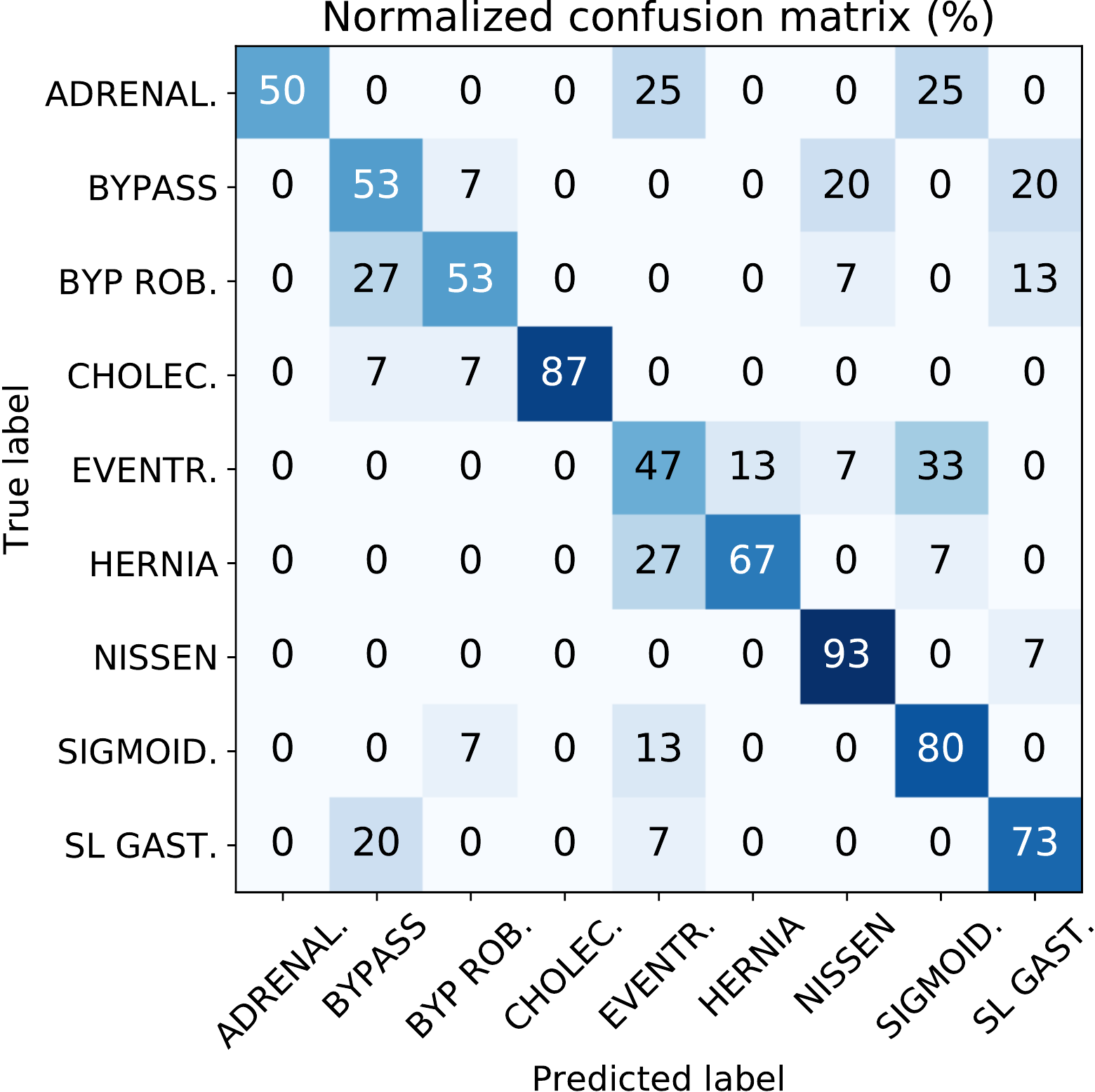}
		\end{subfigure}%
		\begin{subfigure}{0.30\linewidth}
			\centering
			\includegraphics[width=\linewidth]{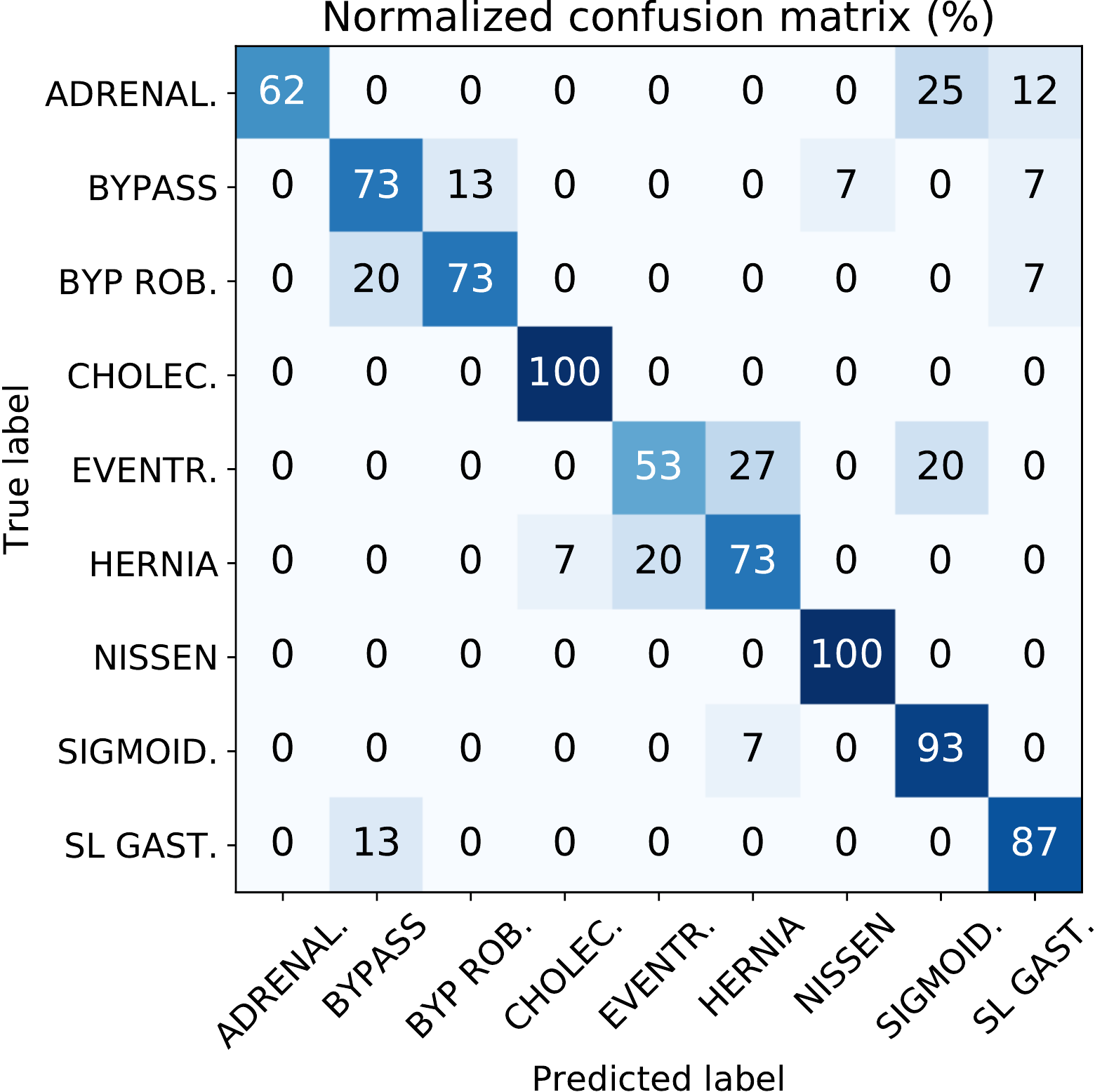}
		\end{subfigure}%
	\caption{Class-wise accuracy (\%) of {FSP-LSTM}  with $\Delta$=10min and $\lambda$=100 after 2min, 8min and 14min (from left to right)}
	\end{subfigure}
	\caption{Confusion matrix for the {\ours}}
	\label{fig:conf_mat}
\end{figure*}

\begin{figure*}[h]
	\begin{subfigure}{0.49\linewidth}
		\includegraphics[width=\linewidth]{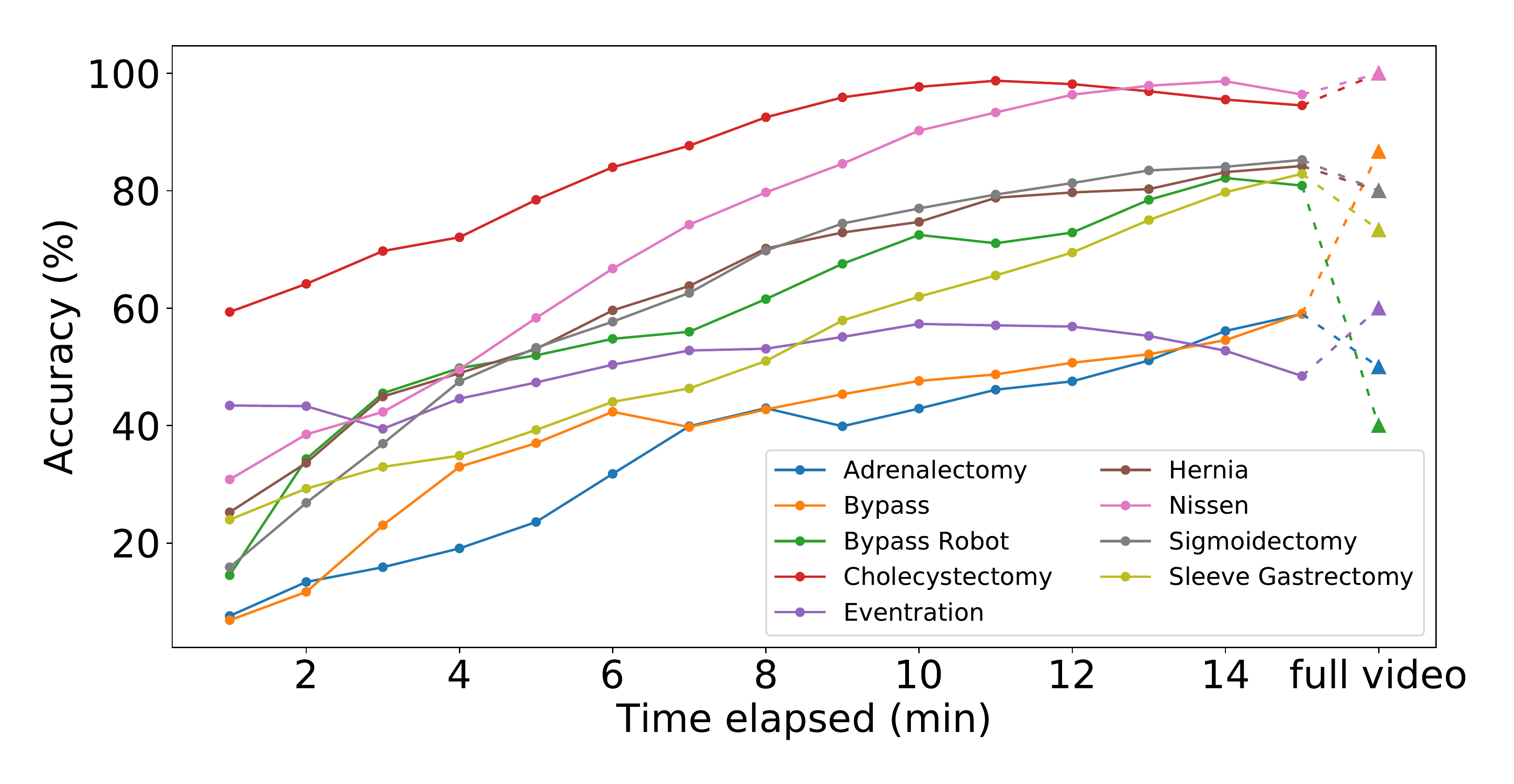}
    	\caption{$\Delta$=0.2 and $\lambda$=10}
	\end{subfigure}
	\begin{subfigure}{0.49\linewidth}
	    \centering
		\includegraphics[width=\linewidth]{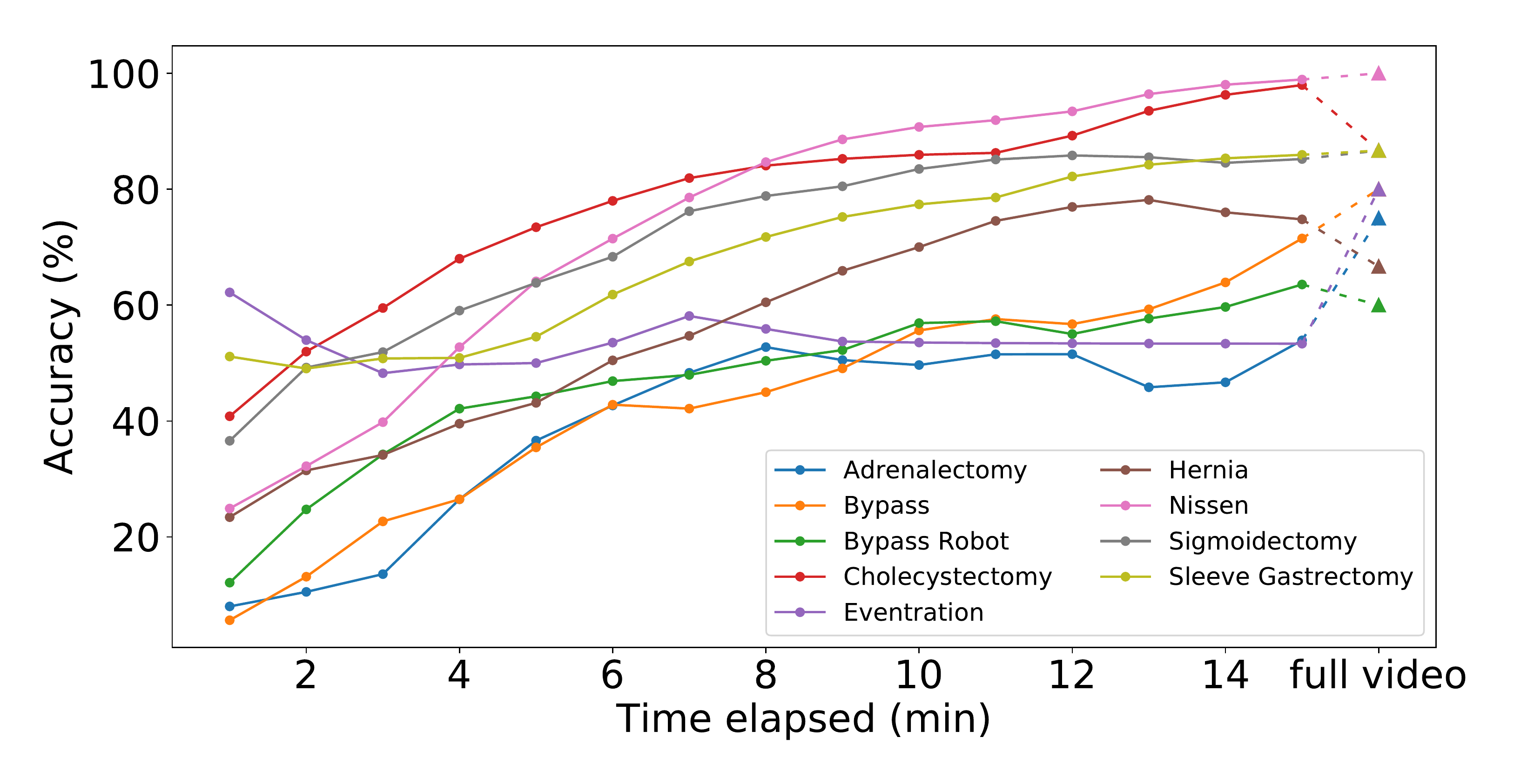}
    	\caption{$\Delta$=10min and $\lambda$=100}
	\end{subfigure}
	\caption{Class-wise early recognition performance  of \ours. The points at the 'full video' mark on the x-axis indicate accuracies when the model has seen the complete video as described in section \ref{sec:cls_wise}.}
	\label{fig:cls_wise_acc_vs_time}
\end{figure*}

\subsection{Class-wise analysis}
\label{sec:cls_wise}
We present in Fig. \ref{fig:conf_mat} the confusion matrices over three time steps and in Fig. \ref{fig:cls_wise_acc_vs_time} the prediction accuracies over time per class. We observe a nice trend wherein the accuracy of prediction increases with time for most surgeries. In fact, both the variants of the FSP-LSTM are able to predict Cholecystectomy and Nissen with nearly 90\% accuracy after 8 minutes. On the other hand, we notice that the recognition performance of Eventration does not improve with time. This happens because Eventration is very often confused with Hernia and Sigmoidectomy. It is interesting to note here that Eventration and Hernia are indeed similar conditions. 

An unexpected observation is the confusion between Bypass and Bypass Robot, which one may have expected the model to resolve by picking up the distinct features of the robotic arm in Bypass Robot. 
While it appears in Fig. \ref{fig:conf_mat} that the initial confusion is gradually getting resolved as time elapses, we found that the  model often misclassifies Bypass Robot as Bypass till the end of the surgeries. This is the reason for the drop in Bypass Robot accuracies at the 'full video' point in Fig. \ref{fig:cls_wise_acc_vs_time}. We suspect that this confusion is because of similar surgical workflows and operating region. We also observed that it is mostly the tool-tips which are present in the video frames and not the full robotic arm, possibly making it difficult to distinguish from regular tools.

The relatively low accuracy of Adrenalectomy can be attributed to the fact that the class has only half the number of videos compared to the other classes to train and test on. We also see that the precision for Sigmoidectomy is low, though the recall is high, in both variants of our model. This could be the effect of our early recognition loss from eqn. \ref{eq:lin_wt_avg_loss} and the fact that the duration of surgeries belonging to different classes is unbalanced. If Sigmoidectomy is similar to other classes in the early stages of the surgery, our model is trained in such a way that the penalty on a false positive for a long video at a given instant is less than that for a shorter video, which can be seen from eqn. \ref{eq:lin_wt_avg_loss}. This explains the tendency to favor Sigmoidectomy over other similar classes that are shorter in duration.

We also studied the performance of our model when it has seen the complete video and compare it with the early recognition performance. This helps us understand the impact of using only the first 15 min. of the surgery as opposed to using the whole surgery. The points at the 'full video' mark in Fig. \ref{fig:cls_wise_acc_vs_time} represent the class-wise accuracy of the model predictions after it has seen each video in the given class completely. It is to be noted here that this accuracy is slightly different from the ones reported at other points on the x-axis in that it does not correspond to a specific duration of time from the beginning of the surgery. This is because the model has to see each video completely and our videos are of variable length. A potential caveat in this comparison is that the model would have observed every phase of a surgery before making a prediction at the 'full video' points. But at other points, different videos may be at different phases when the model makes a prediction. This is not an apples-to-apples comparison but nevertheless serves to get an idea of how close the early recognition performance is to that at the 'full video' point. In Fig. \ref{fig:cls_wise_acc_vs_time} we observe a variable trend at the 'full video' point for different classes. Some classes improve and even touch 100\%, while others drop. It is also to be noted that our models were not originally trained for using the last parts of the video because we use future-state prediction and the last $\Delta$ frames of the video have no future-states. So it is not very clear as to what the model predicts as future state for these last frames at test time but it is interesting to see that accuracy does improve for some classes all the same.

\section{Conclusions} \label{sec:conclusion}
This paper addresses the task of early surgery recognition from laparoscopic videos, a new problem that is critical to ease the real-time deployment of context-aware systems within the OR. Novel methods have been presented for improving early recognition performance, namely the weighted sub-video CNN fine-tuning and {\ours} approaches. We obtain an accuracy of 75\% after having observed only 10 minutes of the surgery, an improvement greater than 10\% over a baseline method, which does not utilize the proposed CNN fine-tuning and future prediction approaches. We believe that 10 minutes is a reasonable time to make a prediction about surgery type given that the average length of surgeries in our dataset is nearly 90 minutes and that most of the critical phases in a surgery occur after this. Thus, the results of our work are highly encouraging from a practical standpoint. In the future, it would be very interesting to explore the potential benefits of the proposed approach in other areas of computer vision as well.

\bibliographystyle{IEEEtran}
\bibliography{fsp}

\begin{thebibliography}{10}
\providecommand{\url}[1]{#1}
\csname url@samestyle\endcsname
\providecommand{\newblock}{\relax}
\providecommand{\bibinfo}[2]{#2}
\providecommand{\BIBentrySTDinterwordspacing}{\spaceskip=0pt\relax}
\providecommand{\BIBentryALTinterwordstretchfactor}{4}
\providecommand{\BIBentryALTinterwordspacing}{\spaceskip=\fontdimen2\font plus
\BIBentryALTinterwordstretchfactor\fontdimen3\font minus
  \fontdimen4\font\relax}
\providecommand{\BIBforeignlanguage}[2]{{%
\expandafter\ifx\csname l@#1\endcsname\relax
\typeout{** WARNING: IEEEtran.bst: No hyphenation pattern has been}%
\typeout{** loaded for the language `#1'. Using the pattern for}%
\typeout{** the default language instead.}%
\else
\language=\csname l@#1\endcsname
\fi
#2}}
\providecommand{\BIBdecl}{\relax}
\BIBdecl

\bibitem{yengera2018}
G.~Yengera, D.~Mutter, J.~Marescaux, and N.~Padoy, ``Less is more: Surgical
  phase recognition with less annotations through self-supervised pre-training
  of cnn-lstm networks,'' \emph{ArXiv}, vol. abs/1805.08569, 2018.

\bibitem{twinanda2018rsd}
A.~P. Twinanda, G.~Yengera, D.~Mutter, J.~Marescaux, and N.~Padoy, ``{RSDNet}:
  Learning to predict remaining surgery duration from laparoscopic videos
  without manual annotations,'' \emph{Arxiv}, vol. abs/1802.03243, 2018.

\bibitem{hajj2017}
H.~A. Hajj, M.~Lamard, P.~Conze, B.~Cochener, and G.~Quellec, ``Monitoring tool
  usage in cataract surgery videos using boosted convolutional and recurrent
  neural networks,'' \emph{CoRR}, vol. abs/1710.01559, 2017.

\bibitem{twinanda2014}
A.~P. Twinanda, J.~Marescaux, M.~De~Mathelin, and N.~Padoy, ``Towards better
  laparoscopic video database organization by automatic surgery
  classification,'' in \emph{Information Processing in Computer-Assisted
  Interventions}.\hskip 1em plus 0.5em minus 0.4em\relax Cham: Springer
  International Publishing, 2014, pp. 186--195.

\bibitem{aliakbarian2017iccv}
M.~Sadegh~Aliakbarian, F.~Sadat~Saleh, M.~Salzmann, B.~Fernando, L.~Petersson,
  and L.~Andersson, ``Encouraging lstms to anticipate actions very early,'' in
  \emph{The IEEE International Conference on Computer Vision (ICCV)}, Oct 2017.

\bibitem{laptev2008cvpr}
I.~Laptev, M.~Marszalek, C.~Schmid, and B.~Rozenfeld, ``Learning realistic
  human actions from movies,'' in \emph{2008 IEEE Conference on Computer Vision
  and Pattern Recognition}, June 2008, pp. 1--8.

\bibitem{wang2013iccv}
H.~Wang and C.~Schmid, ``Action recognition with improved trajectories,'' in
  \emph{2013 IEEE International Conference on Computer Vision}, Dec 2013, pp.
  3551--3558.

\bibitem{karpathy2012}
A.~Krizhevsky, I.~Sutskever, and G.~E. Hinton, ``Imagenet classification with
  deep convolutional neural networks,'' in \emph{Advances in Neural Information
  Processing Systems 25}.\hskip 1em plus 0.5em minus 0.4em\relax Curran
  Associates, Inc., 2012, pp. 1097--1105.

\bibitem{simonyan_vgg14}
K.~Simonyan and A.~Zisserman, ``Very deep convolutional networks for
  large-scale image recognition,'' \emph{CoRR}, vol. abs/1409.1556, 2014.

\bibitem{he2016}
K.~He, X.~Zhang, S.~Ren, and J.~Sun, ``Deep residual learning for image
  recognition,'' in \emph{2016 IEEE Conference on Computer Vision and Pattern
  Recognition (CVPR)}, June 2016, pp. 770--778.

\bibitem{simonyan2014}
K.~Simonyan and A.~Zisserman, ``Two-stream convolutional networks for action
  recognition in videos,'' in \emph{Advances in Neural Information Processing
  Systems 27}.\hskip 1em plus 0.5em minus 0.4em\relax Curran Associates, Inc.,
  2014, pp. 568--576.

\bibitem{ng2015}
J.~Y.-H. Ng, M.~J. Hausknecht, S.~Vijayanarasimhan, O.~Vinyals, R.~Monga, and
  G.~Toderici, ``Beyond short snippets: Deep networks for video
  classification.'' in \emph{CVPR}.\hskip 1em plus 0.5em minus 0.4em\relax IEEE
  Computer Society, 2015, pp. 4694--4702.

\bibitem{hochreiter1997long}
S.~Hochreiter and J.~Schmidhuber, ``Long short-term memory,'' \emph{Neural
  computation}, vol.~9, no.~8, pp. 1735--1780, 1997.

\bibitem{twinanda2017endonet}
A.~P. Twinanda, S.~Shehata, D.~Mutter, J.~Marescaux, M.~de~Mathelin, and
  N.~Padoy, ``Endonet: A deep architecture for recognition tasks on
  laparoscopic videos,'' \emph{IEEE Transactions on Medical Imaging}, vol.~36,
  no.~1, pp. 86--97, 2017.

\bibitem{zappella2012miccai}
B.~B{\'e}jar~Haro, L.~Zappella, and R.~Vidal, ``Surgical gesture classification
  from video data,'' in \emph{Medical Image Computing and Computer-Assisted
  Intervention -- MICCAI 2012}, N.~Ayache, H.~Delingette, P.~Golland, and
  K.~Mori, Eds.\hskip 1em plus 0.5em minus 0.4em\relax Berlin, Heidelberg:
  Springer Berlin Heidelberg, 2012, pp. 34--41.

\bibitem{reiter2012miccai}
A.~Reiter, P.~K. Allen, and T.~Zhao, ``Feature classification for tracking
  articulated surgical tools,'' in \emph{Medical Image Computing and
  Computer-Assisted Intervention -- MICCAI 2012}, N.~Ayache, H.~Delingette,
  P.~Golland, and K.~Mori, Eds.\hskip 1em plus 0.5em minus 0.4em\relax Berlin,
  Heidelberg: Springer Berlin Heidelberg, 2012, pp. 592--600.

\bibitem{petscharnig2018}
\BIBentryALTinterwordspacing
S.~Petscharnig and K.~Sch{\"o}ffmann, ``Learning laparoscopic video shot
  classification for gynecological surgery,'' \emph{Multimedia Tools and
  Applications}, vol.~77, no.~7, pp. 8061--8079, Apr 2018. [Online]. Available:
  \url{https://doi.org/10.1007/s11042-017-4699-5}
\BIBentrySTDinterwordspacing

\bibitem{varma2009}
\BIBentryALTinterwordspacing
M.~Varma and B.~R. Babu, ``More generality in efficient multiple kernel
  learning,'' in \emph{Proceedings of the 26th Annual International Conference
  on Machine Learning}, ser. ICML '09.\hskip 1em plus 0.5em minus 0.4em\relax
  New York, NY, USA: ACM, 2009, pp. 1065--1072. [Online]. Available:
  \url{http://doi.acm.org/10.1145/1553374.1553510}
\BIBentrySTDinterwordspacing

\bibitem{jain2016}
A.~Jain, A.~Singh, H.~S. Koppula, S.~Soh, and A.~Saxena, ``Recurrent neural
  networks for driver activity anticipation via sensory-fusion architecture,''
  in \emph{2016 IEEE International Conference on Robotics and Automation
  (ICRA)}, May 2016, pp. 3118--3125.

\bibitem{ranzato2014}
\BIBentryALTinterwordspacing
M.~Ranzato, A.~Szlam, J.~Bruna, M.~Mathieu, R.~Collobert, and S.~Chopra,
  ``Video (language) modeling: a baseline for generative models of natural
  videos,'' \emph{CoRR}, vol. abs/1412.6604, 2014. [Online]. Available:
  \url{http://arxiv.org/abs/1412.6604}
\BIBentrySTDinterwordspacing

\bibitem{srivastava2015}
\BIBentryALTinterwordspacing
N.~Srivastava, E.~Mansimov, and R.~Salakhudinov, ``Unsupervised learning of
  video representations using lstms,'' in \emph{Proceedings of the 32nd
  International Conference on Machine Learning}, ser. Proceedings of Machine
  Learning Research, F.~Bach and D.~Blei, Eds., vol.~37.\hskip 1em plus 0.5em
  minus 0.4em\relax Lille, France: PMLR, 07--09 Jul 2015, pp. 843--852.
  [Online]. Available: \url{http://proceedings.mlr.press/v37/srivastava15.html}
\BIBentrySTDinterwordspacing

\bibitem{lotter2017}
W.~Lotter, G.~Kreiman, and D.~Cox, ``Deep predictive coding networks for video
  prediction and unsupervised learning,'' in \emph{ICLR}, 2017.

\bibitem{vondrick2016}
C.~Vondrick, H.~Pirsiavash, and A.~Torralba, ``Anticipating visual
  representations from unlabeled video,'' in \emph{The IEEE Conference on
  Computer Vision and Pattern Recognition (CVPR)}, June 2016.

\bibitem{villegas17}
R.~Villegas, J.~Yang, Y.~Zou, S.~Sohn, X.~Lin, and H.~Lee, ``{Learning to
  Generate Long-term Future via Hierarchical Prediction},'' in \emph{ICML},
  2017.

\bibitem{wang2016}
L.~Wang, Y.~Xiong, Z.~Wang, Y.~Qiao, D.~Lin, X.~Tang, and L.~{Val Gool},
  ``Temporal segment networks: Towards good practices for deep action
  recognition,'' in \emph{ECCV}, 2016.

\bibitem{chen2017}
Q.~Chen and Y.~Zhang, ``Sequential segment networks for action recognition,''
  \emph{IEEE Signal Processing Letters}, vol.~24, no.~5, pp. 712--716, May
  2017.

\end{thebibliography}

\end{document}